\documentclass[a4paper,fleqn]{cas-sc}

\usepackage[numbers]{natbib}
\usepackage{float}
\usepackage{caption}
\usepackage[tableposition=top]{caption}
\usepackage[justification=centering]{caption}
\usepackage{cleveref}
\floatstyle{plaintop}
\restylefloat{table}
\usepackage{graphics}
\usepackage{rotating}
\usepackage{soul}
\usepackage{longtable}
\usepackage{multicol}
\usepackage{acronym}
\usepackage{subcaption}
\usepackage{tabu}
\usepackage[labelfont=bf, skip=5pt, font=small]{caption}
\usepackage{graphicx}
\usepackage{fancyhdr}
\usepackage{multirow}

\usepackage{amsmath}
\usepackage{amsthm}
\usepackage{amssymb}

\usepackage{graphicx}

\def\tsc#1{\csdef{#1}{\textsc{\lowercase{#1}}\xspace}}
\tsc{WGM}
\tsc{QE}
\tsc{EP}
\tsc{PMS}
\tsc{BEC}
\tsc{DE}
\DeclareUnicodeCharacter{2212}{-}
\begin{document}
\let\WriteBookmarks\relax
\def\floatpagepagefraction{1}
\def\textpagefraction{.001}
\shorttitle{}
\shortauthors{Ogunfowora et al.}

\title [mode = title]{A Transformer-based Framework For Multi-variate Time Series: A Remaining Useful Life Prediction Use Case.} 

\author[1]{Oluwaseyi Ogunfowora}[type=editor,
                        auid=000,bioid=1]
\ead{ogunfool@uvic.ca}
\credit{Conceptualization, Methodology, Formal analysis, Investigation, Data Curation, Visualization, Writing - Original Draft, Writing - Review \& Editing}

% \author[1]{Mat Harisson\corref{correspondingauthor}}
\author[1,2]{Homayoun Najjaran\corref{correspondingauthor}}
% \cormark[]
\ead{najjaran@uvic.ca}
\credit{Writing - Review \& Editing, Supervision} 

\address[1]{Department of Mechanical Engineering, University of Victoria, Victoria BC, V8P 5C2, Canada}
\address[2]{Department of Electrical and Computer Engineering, University of Victoria, Victoria BC, V8P 5C2, Canada}
\cortext[correspondingauthor]{Corresponding author}
% \cortext[cor1]{Corresponding author}

\begin{abstract}
In recent times, large language models (LLMs) have captured a global spotlight and revolutionized the field of Natural Language Processing. One of the factors attributed to the effectiveness of LLMs is the model architecture used for training, transformers. Transformer models excel at capturing contextual features in sequential data since time series data are sequential, transformer models can be leveraged for more efficient time series data prediction.\\
The field of prognostics is vital to system health management and proper maintenance planning. A reliable estimation of the remaining useful life (RUL) of machines holds the potential for substantial cost savings. This includes avoiding abrupt machine failures, maximizing equipment usage, and serving as a decision-support system (DSS). Data-driven methods for predictive maintenance have been recognized as one of the most promising maintenance strategies because of their high efficiency and low cost compared to other strategies.\\
This work proposed an encoder-transformer architecture-based framework for multi-variate time series prediction with a prognostics use case. We validated the effectiveness of the proposed framework on all four sets of the C-MAPPS benchmark dataset for the remaining useful life prediction task. To effectively transfer the knowledge and application of transformers from the natural language domain to time series, three model-specific experiments were conducted. Layer normalization versus batch normalization layers, fixed versus learnable positional encodings, and three input data transformation methods were experimented with. Also, a clustering-based normalization technique was developed to pre-process the data and extract relevant features. To enforce weak-sense stationarity in the time series data and enable the model's awareness of the initial stages of the machine life and its degradation path, a novel expanding window method was proposed for the first time in this work, it was compared with the sliding window method and it led to a large improvement in the performance of the encoder-transformer model.\\
Finally, the performance of the proposed encoder-transformer model was evaluated on the test dataset and compared with the results from 13 other state-of-the-art (SOTA) models in the literature and it outperformed them all with an average performance increase of 137.65\% over the next best model across all the datasets.
\end{abstract}


\maketitle
% \footnote[1]{We share our code for the proposed end-to-end RUL prediction workflow at: https://github.com/Ogunfool}
\section{Introduction}
\label{Time series data and The Remaining Useful Life Prediction Problem}
\subsection{Time series data and The Remaining Useful Life Prediction Problem}
% \textbf{Time series Data Prediction}\\
Time series data is a sequence of measurements taken at successive equally spaced intervals over time. Time series data prediction task finds utility across diverse domains and applications, including but not limited to predictive maintenance, manufacturing, bio-signal processing, finance, meteorology, and more.\\
Time series data prediction and forecasting is one of the foremost domain applications
of predictive modeling. A variety of modeling approaches have been developed
in the literature for univariate and multivariate time series data. Statistical methods
such as auto-regressive models and their variates (ARIMA, VARMA, VARIMA, etc) or deep, and non-deep learning ML methods such as TS-CHIEF [59] have been used in the literature. Nevertheless, unlike computer vision or Natural Language Processing (NLP) domains, deep learning models have not yet achieved supremacy or yielded unmatched outcomes when compared to non-deep learning models in time series domain applications.
According to \cite{t2}, before the inception of transformer models, non-deep learning methods such as TS-CHIEF \cite{t1}, HIVE-COTE \cite{t3}, and ROCKET \cite{t4} held the record on time series regression and classification dataset benchmarks, closely matching or even outperforming sophisticated deep learning architectures like InceptionTime and ResNet. \\
Time series data prediction is a non-trivial task and machine learning models might find it hard to make accurate predictions because of the following reasons:
\begin{enumerate}
\item \text Machine learning models were developed under the assumption that the data points are independent and identically distributed random variables, but time series data does not obey this assumption. Time series data are not mutually independent, there is auto-correlation between the data points. Also, they are not identically distributed random variables, as the name implies time series data changes over time, and the distribution (mean and standard deviation) changes over time, therefore they are non-stationary. While different methods such as differencing, detrending, or finding the time interval where weak-sense stationarity can be observed in the data, have been developed in the literature to make time series data more stationary, these underlying characteristics of time series data make it difficult for machine learning models to make very good predictions.
\item \text Other distinct characteristics particular to time series data are trends, seasonality, cycles, special events, or a combination of all these characteristics. Changes in the trends or seasonality patterns in the data make time series data very difficult to predict. These distinctive traits observed in univariate time series data even become more difficult to handle in multi-variate time series data.
\item \text Finally, unlike language, which is universal, and has led to the widespread success of Natural Language Processing (NLP) because of their ability to learn encoded representations of natural language through unsupervised learning approaches, time series data comes in different kinds and forms, they do not have a universal structure. This makes it difficult to generalize the encoded representation of a particular data/use case to another data/use case.\\
Multi-dimensional time series modeling is difficult and sophisticated models are required to make better predictions. 
Since time series data can be modeled as sequential data, it is only reasonable to transfer the concept of how sequential data has been modeled in other domains like natural language processing to the time series data prediction and modeling domain. The base architecture for large language models is \textbf{transformers} and this work adopts it for better time-series data predictions.\\
\end{enumerate}

\textbf{The Remaining Useful Life Prediction Task}\\
Maintenance activities take up 15\%-40\% of the total production costs in factories \cite{r1}. Machines/assets undergo various failure modes that result in machine health degradation, affect system performance, and eventually cause machine failure. Degradation of machines whether under working or non-working conditions is inevitable and so is the need for maintenance.
While the cost associated with maintenance activities cannot be totally eliminated, a proper maintenance plan can help to minimize these costs.\\
For most industries to remain competitive, they cannot afford the short or long-term costs and effects associated with inadequate planning of production and maintenance activities which can result in not meeting customer demands and loss of sales. \cite{ogunfowora2023reinforcement}\\
The field of prognostics is vital to systems health management and proper maintenance planning. The reliable estimation of remaining useful life (RUL) holds the potential for substantial cost savings. This includes avoiding unscheduled maintenance and maximizing equipment usage, serving as decision-support systems (DSS) by providing decision-makers valuable information about the condition of the machine and helping them to plan accordingly. For instance, this can inform their decision to reduce the operational loads on the machines in order to extend their life span, these estimations can also enable planners to anticipate upcoming maintenance needs and initiate a seamless logistics process, facilitating a smooth transition from faulty equipment to fully functional ones.\\
Predictive maintenance has been recognized as one of the most promising maintenance strategies for production systems because of its high efficiency and low cost compared to other strategies \cite{r4}. This approach helps to eliminate the unnecessary maintenance costs incurred in the scheduled maintenance approach while significantly reducing unscheduled breakdowns because the maintenance decisions are made based on the changing, real-time machine health conditions. Numerous works have been done in literature to harness the capabilities of machine-learning-based predictive models to accurately predict machine failures and make diagnoses of the failure types.\\
The remaining useful life (RUL) prediction problem at any time during the machine's lifetime can be modeled as a regression task by representing the target as the actual time-to-failure values. That is, depending on how the successive data points are collected in time, given that we are at timestep $t$ if the machine will fail at time step $t + 10$, then the target (RUL) at timestep $t$ will be $(t + 10) − t.$\\

\textbf{Transformers for time series}\\ 
The base model used in LLM models such as BERT and GPT is \textbf{transformers}. Transformers are highly expressive models that are excellent for sequential data modeling, they are built upon the foundation of \textbf{attention} mechanisms. Attention is a resource allocation mechanism that mimics the brain by recognizing what or where to pay more attention to and assigning larger weights to them. Attention mechanisms were developed to solve the problem of long-term dependencies experienced in recurrent neural networks due to very long data sequences.\\  Since time series data is sequential data, we can use sequential modeling approaches like transformers and attention mechanisms to model it.\\
Contextual features can help make better predictions on time series data. Usually, these contextual features are manually or artificially generated. Transformers have the ability to extract these contextual features on their own due to their inherent ability to capture contextual relationships within data, especially in sequences.
Transformers excel at extracting contextual features because of their self-attention mechanism, which allows them to weigh the importance of different elements in a sequence based on their relationships with other elements.
\subsection{Contributions of this work.}
\begin{enumerate}
    \item An end-to-end RUL prediction framework for multi-variate time series degradation datasets with multiple operating conditions and failure modes is developed based on the encoder-transformer architecture.\\
    To the best of our knowledge, we present for the first time a native encoder-transformer architecture for the RUL prediction tasks on the C-MAPPS benchmark dataset that has shown a higher and/or competitive performance over other augmented transformer architectures that have been used previously for this task in the literature.
    Augmented transformer architectures in this context refer to transformer architectures that differ from the originally proposed architecture in the "Attention is All you Need" paper \cite{t6} based solely on the self-attention mechanism and dispensing the recurrent and convolutional neural networks entirely.\\
    Due to the difficulty in time-series data prediction tasks and the specificity of this application, augmented transformers or attention-based architectures combined with recurrent, convolution networks and even channel-attention-based transformer models have been commonly used for the RUL prediction task. Some of these models are, the use of a bi-directional gated recurrent unit with the transformer encoder architecture in \cite{t22}, and \cite{t11} used an encoder-decoder transformer network combined with CNN-based channel attention for feature extraction.\\
    This work claims that the native transformer architectures used for LLM(s) based solely on the self-attention mechanism can be used to achieve competitive results for the RUL prediction task on the NASA C-MAPPS datasets. 
    \item Following the work of \cite{t2} we investigate the proposed changes that can be made to the original transformer architecture to make it compatible with multivariate time series data prediction task, instead of sequences of discrete word indices.\\
    In the “Attention is All you Need” paper \cite{t6}, where transformers based solely on self-attention mechanisms were first introduced; the experiments were performed on machine translation tasks. However, in this work, we aim to find ways to effectively leverage transformer models to make better predictions on time series data. To achieve this and transfer knowledge from the NLP domain, it is expedient that while following the base transformer architecture like using a self-attention mechanism and skip connections, we also need to tailor some of the transformer modules for better time series (RUL) prediction tasks. \\
    Three model-specific experiments were conducted, layer normalization versus batch normalization layers, fixed versus learnable positional encodings, and various input data transformation methods were experimented with for the RUL prediction task. The result of these experiments validates some of the claims and recommendations of the authors of \cite{t2} which were experimented on popular benchmark time series prediction and classification tasks.\\
    This work further generalizes these claims by experimenting with the NASA turbo-engine run-to-failure degradation time series datasets with unique characteristics.
    \item Finally, a novel method of data preparation of time series data is proposed. The expanding window method is developed and compared with the sliding window method. The expanding window method resulted in better prediction performance of the transformer model than the sliding window method of constant length. \\
    This palpable improvement in the performance of the expanding window method over the sliding window method was attributed to the degradation paths of machines which can be greatly influenced by the initial stages of the machine's life. This helps the model to be aware of the early failure stages and changing degradation paths of the machines.
\end{enumerate}

\subsection{Organizational Structure / Workflow}
\begin{description}
\item[\textbf{Chapter 1}] introduces the various applications of time-series data prediction tasks which include the remaining useful life prediction task performed in this work. The importance of the prognostics and health management sub-field of predictive maintenance is also introduced. In this chapter, emphasis is laid on why accurate predictions of time series data are difficult, the remaining useful life prediction problem is explained and transformers model suitability for time series data prediction tasks are discussed.\\
Finally, the contributions of this work to literature are presented.
\item[\textbf{Chapter 2}] contains the related works, the two main machine-learning-based RUL prediction methods, the similarity-based model (SBM) and direct approximation method are briefly introduced. Other works in the literature that used the direct approximation method for RUL prediction tasks are briefly discussed.
\item[\textbf{Chapter 3}] focuses on the attention mechanism and transformer architecture. It describes the different modules and layers that make up a single transformer block. The attention-based models and mechanism, the transformer model which includes the self-attention, multi-head attention, positional encodings, layer normalization methods and skip connections, and the fully connected layers are discussed.\\
At the end of this chapter, the case study dataset and the evaluation metrics used in this work are described.
\item[\textbf{Chapter 4}] describes the proposed RUL prediction framework in detail. It presents and shows the experimental results of the model-specific and application-specific experiments conducted. A workflow of the final RUL prediction method used in this work is presented and developed for the datasets.  
\item[\textbf{Chapter 5}] presents the results of the proposed models, it includes the evaluation on the reference datasets and the comparisons with other state-of-the-art models in the literature.
\item[\textbf{Chapter 6}] contains the conclusion, a restatement of the claims, and the areas of future work.
\end{description}

% \newpage
\section{Related Work} \label{sec:biblio}
In the literature, the two main ML-based methods for RUL prediction are similarity-based and direct approximation methods. Similarity-based models (SBM) use the similarity in the degradation profile of the training data; unique run-to-intervention instances from similar components/machines under similar operating conditions (ensemble members) to estimate the RUL of the test component. The two main stages of development of an SBM are the health index (HI) construction stage and the similarity measurement and RUL fusion stage. \\Similarity-based models gained popularity from the first Prognostics and Health Management competition held in 2008. The winners of the competition, \cite{sb1} used a similarity-based model, and ever since researchers have continually found ways to improve the performance of SBMs.
Even though the similarity-based model won the PHM08 competition, direct approximation models came in as the first and second runner-ups. Extensive research has also been done in improving the performance of the direct approximation methods. \\

Direct approximation methods refer to using machine learning models to directly predict the remaining useful life of a machine at any time during its life. The ML models serve as function approximators that learn the relationship between the input and the target which in this case is the remaining useful life.\\
Typically, traditional machine learning techniques require effective feature engineering, encompassing tasks such as feature extraction and dimensionality reduction. In the absence of pertinent human knowledge, the utilization of improper features tends to lead to sub-optimal performance.
More generally, in the direct approximation methods research works for RUL prediction, more attention has been paid to deep learning methods because of their complexity, expressiveness, and ability to learn non-linear and complex features in data. More particularly convolutional neural networks (CNNs) and recurrent neural networks have shown impressive results on time series data. \\
Convolutional NNs are a type of deep-learning neural network that uses convolutions to find patterns in data. The idea behind convolution is to use kernels to extract features from input data, the concept of using kernels for an image or signal processing is not new. In the past, kernels of known weights e.g., the Gaussian Blur Filter were used to blur images or used on time series data for smoothing by reducing the level of noise in them. What makes CNNs special is the fact that they use learnable kernels which is a common characteristic of deep neural networks, the features are learned. While CNNs have been primarily used to solve image-driven pattern recognition tasks, they have also shown great potential in their applications to time series data prediction. This is because we can use the auto-correlations in time series data to enhance the predictive abilities which is the fundamental building block of auto-regressive (AR(p)) models (linear models), so autoregressive CNNs can be thought of as an extension of AR(p) models.\\ In \cite{t13} and \cite{t14} deep CNNs were used for remaining useful life prediction. \cite{t10} established a multi-scale CNN with a powerful feature extraction ability.\\

RNNs are one of the most used deep learning models for time series data modeling. The output of RNNs at a given time step $t$, is not only influenced by the input at that time step but also by the hidden state representation which contains a memory of all the states that have been visited before the current state. This is why RNNs are said to have memory.
The vanilla RNNs faced the issue of vanishing gradients which hindered their learning and performance; long-short-term memory (LSTM(s)) and gated recurrent units (GRUs) that use gates to decide how much of the hidden representation from previous layers should be remembered in the current time step were developed to address this issue. \\
Several scholars have conducted research on
RUL estimation using LSTM networks, \cite{rnn1}, \cite{rnn2}, and \cite{rnn3} used LSTM neural networks for RUL prediction tasks. \cite{rnn4} took it further by using bidirectional LSTM (Bi-LSTM)-based networks for predicting RUL. In contrast to previous studies that only used LSTM in a forward orientation, the bi-directional LSTM architecture processes input sequences in both forward and backward directions. It consists of two separate LSTM layers: one processes the input sequence in a forward direction, while the other processes it in a reverse (backward) direction. This enables the model to capture contextual information from both past and future elements of the input sequence simultaneously.\\

 \cite{t16} and \cite{t19} used a combination of fully connected neural networks and LSTMs for RUL predictions. Particularly interesting is the RBM-LSTM-FNN RUL prediction model developed by \cite{t18}, this model uses a semi-supervised approach to improve the prediction accuracy. The authors investigated the effect of unsupervised pre-training in RUL predictions utilizing a semi-supervised setup. They also used a meta-heuristic algorithm, the Genetic Algorithm (GA) approach to tune the large hyperparameter space.\\
CNNs and RNNs have also been combined by some authors, \cite{liu2019novel} combined a Bi-LSTM and CNN to identify temporal relationships and essential features from time series, and the fully connected layer was used in the final layer.\\

In sequential data modeling, resulting from the need for very long sequences in the fields of Natural Language Processing (NLP) and Large Language Models (LLMs) development, a common challenge encountered is the problem of long-term dependencies. As the input data sequence lengths increase, all the problems associated with training very deep neural networks like vanishing or exploding gradients which leads to non-convergence come to play. Attention-based networks that use attention mechanisms were introduced to improve the performance of sequence-to-sequence models like RNNs. The attention mechanisms gave rise to the self-attention mechanism which is the fundamental building block of a transformer network.\\
Since time series data can be modeled as sequential data, researchers have also adopted attention-based and transformer networks for RUL prediction and these models have shown tremendous capabilities. The authors of \cite{t22} proposed a bi-directional gated recurrent unit with a temporal self-attention mechanism for RUL prediction. In \cite{t23} a transformer encoder architecture with a gated convolutional unit was developed to extract the local features and the encoder transformer to extract the global features. \cite{t11} used an encoder-decoder transformer network with CNN-based channel attention for feature extraction.\\

% \newpage
\section{Attention-based Models and Transformer Architecture} \label{sec:graph}
\textbf{Attention-based Models and Mechanism}\\
Attention-based models are models that use attention mechanisms. The attention mechanism allows the base models, usually neural networks (ANN, CNN, or RNNs) to focus on the most important features of the input that produces better results at the output. The presence of this attention mechanism helps to improve the performance of the base models. Attention-based models have gained more traction in sequence-to-sequence models like recurrent neural networks (RNNs).\\
RNNs are one of the most used deep learning models for sequential data modeling. The output of RNNs at a given time step $t$, is not influenced only by the input at that time step but also by the hidden state representation which contains a memory of all the states that have been visited before the current state. This is why RNNs are said to have memory.
A common challenge encountered by the naïve RNNs, the Elman units is the problem of long-term dependencies. As the input data sequence lengths increase, the more deeply nested and further back in the sequence a hidden unit is, and due to the compounding multiplicative effect when performing gradient descent and learning the weights, the more likely it is for the gradients to vanish.
This was the issue faced by the vanilla RNN(s); long-short-term memory (LSTM(s)) and gated-recurrent neural networks (GRU(s)) that use gates to decide how much of the hidden representation from previous layers should be remembered in the current time step were developed to fix this issue. However, in most applications, the sequence lengths are required to be very long, and this brings out all the problems associated with training very deep neural networks like vanishing or exploding gradients which leads to non-convergence, so attention networks are used to improve the performance of RNNs.\\

\textbf{Attention Mechanism}\\
To focus on the most important parts of the input to get better outputs, context vectors $(C(t))$ which is a dot product between the hidden representation and learnable weights to all outputs at every time step in the sequence are used. They determine which portions of the hidden representations are important to generate the output at any time instance, $t$. So, at every time step of the decoder, the context vector is different.\\
The attention mechanism was introduced by \cite{t5}. The computation of the attention mechanism is divided into three main steps, the computation of the alignment scores, the weights, and finally the context vector.
\begin{enumerate}
\item \text	Compute alignment scores: The alignment score is a function of the hidden states of the encoder, $h_j$, and the decoder output at $t-1$, $s_{t-1}$. The alignment model can be implemented by a feed-forward neural network. The alignment score as shown in equation \ref{eqn 1} can be generally computed with three methods shown in equation \ref{eqn 2}.
\begin{equation}
    score(s_{t-1}, h_j) = s_{t,j} = \label{eqn 1}\\
\end{equation}
\begin{equation}
 score(s_t,j) =
    \begin{cases}
        (s_t^T)h_j & \text{Dot-product Attention} \label{eqn 2}\\
        (s_t^T)W_ah_j & \text{General Attention}\\
        V_a^Ttanh(W_a[s_t;h_j]) & \text{Additive Attention}\\ 
    \end{cases}
\end{equation}
\item \text	Compute attention weights by applying softmax operation on the alignment scores as seen in equation \ref{eqn 3}.
\begin{equation}
  \alpha_{(s_{t,j})} = \frac{\exp(s_{t,j})}{\sum_j \exp(s_{t,j})}\label{eqn 3}  
\end{equation}
\item \text	Compute context vector: As mentioned above, the context vector is a weighted sum of the attention weights and the hidden state representations for every output. The equation \ref{eqn 4} below shows the computation of the context vector for a single output node.
\begin{equation}
  C_t = \sum_{j=1}^{T} \alpha_{(s_{t,j})}h_j \label{eqn 4}
\end{equation} 

\end{enumerate}
More generally, the attention mechanism can be computed using three main components, the queries Q, the keys K and, the values V. The query can be compared with the decoder output from the previous layer $s_{t-1}$ and the key with the encoder inputs $h_j$ of the Bahdanau attention mechanism.\\

\begin{figure}[h!]
    \centering
    \includegraphics[width=12.5cm]{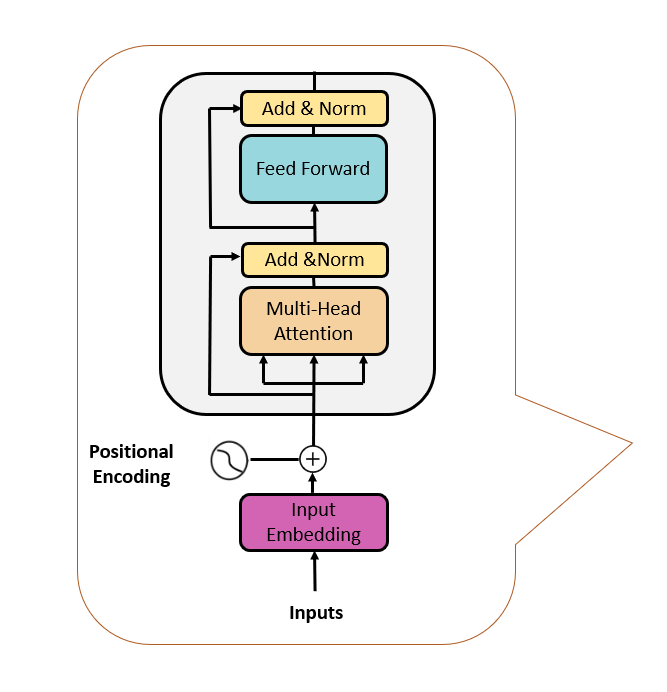}
    \caption{Layers of a single transformer block.}
    \label{fig t2}
\end{figure}

\textbf{Self-attention Mechanism and Transformer Network}\\
The self-attention mechanism is the fundamental building block of a transformer network. The self-attention mechanism finds the similarity between the input sequence. The idea is to create context-aware vectors by transforming the input vectors into their own queries, keys, and values and applying the generalized attention mechanism. Simply put, the queries, keys, and values are computed \textbf{on} the same sequence unlike in the initially introduced attention mechanisms where the query is computed on the decoder values and keys on the encoder values.\\
Self-attention-based networks are superior to RNN(s) or attention-based networks with RNN(s) because they eliminate the need for a base seq2seq model like RNN(s) before computing attention. The attention mechanism can be performed directly on the input data sequence so parallelization is possible while eliminating the long-term dependence problems of RNNs and other problems that arise from long sequences and it can also effectively manage variable length input sequences.\\
A basic transformer block is made up of four main sub-modules which are: multi-head attention (MHA), positional encoding, layer norm and skip connections, and feed-forward neural network layers.  Figure \ref{fig t2} shows the modules that make up a single transformer block.

\begin{enumerate}
\item \textbf{The multi-head attention:} The multi-head attention is concatenated multiple attention heads so that each attention head can pay attention to different parts of the input sequence. Single-head attention can also be computed in two (2) major steps; compute the queries, keys, and values from the input sequence, and use the scaled-dot product attention mechanism. The scaled-dot product is done by computing the dot products of the query with the keys, dividing each by $\sqrt{dk}$, and applying the softmax function to obtain the attention weights.\\
The queries, keys, and values can be computed as:
\begin{equation}
\begin{cases}
  q_i = W^{(Q)T}x_i\\ \label{eqn 5}
  k_i = W^{(K)T}x_i & \text{for $i=1...T$}\\
  v_i = W^{(V)T}x_i \\
\end{cases}
\end{equation} 
where, $q_i$ is the query, $k_i$ the key and $v_i$ the value at a given time step and $W^{(Q)T}$, $W^{(K)T}$, $W^{(V)T}$ are trainable parameters. \\

With $W^{(Q)T}$, $W^{(K)T}$, $W^{(V)T}$, the scaled dot product equation can be computed as,
\begin{equation}
    Attention(Q, K, V) = softmax(\frac{QK^T}{\sqrt{d_k}})V \label{eqn 6}
\end{equation}
Where where $\sqrt{d_k}$ is the dimension of the key vector $k$ and query vector $q$ .\\

Multi-head attention (MHA) can be computed as,
\begin{equation}
    MultiHead(Q, K, V) = Concat(head_1, ..., head_h)W^A \label{eqn 7}
\end{equation}

\item \textbf{Positional Encoding}: As earlier stated, sequential data is a type of data in which order matters. However, after performing self-attention on the input sequence, the sequence is randomly rearranged commonly referred to as \textbf{“permutation invariant”}, so, to keep the data sequence in order, position encodings are used to add positional information to the input.
\item \textbf{Layer normalization and skip connections:} The output of the multi-head attention (MHA) is fed to a layer normalization layer after adding skip connections to the MHA output. The skip connections are used to help the network learn better the same way it was applied in convolutional neural networks like RESNET. It was observed that merely adding more layers to a neural network does not improve the performance instead it degrades it, so, explicitly adding identity layers such that the networks need to learn only residual representations makes the network perform better.
\item \textbf{Fully connected layer:} Many fully connected layers can be stacked together after the MHA block with skip connections and layer norm.
\end{enumerate}

\begin{figure}[h!]
    \centering
    \includegraphics[width=\textwidth]{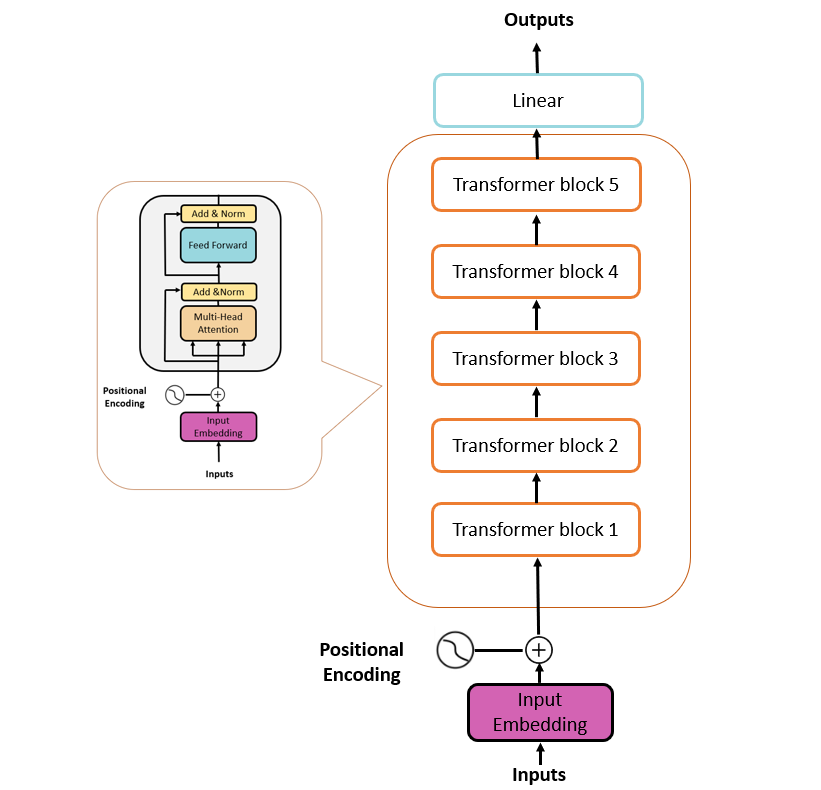}
    \caption{An Encoder-Transformer Architecture.}
    \label{fig t3}
\end{figure}

Finally, three (3) main architectures can be developed from basic transformer blocks. Just like in other seq2seq models, these architectures are generally referred to as the encoder, decoder, and encoder-decoder architectures.\\
Unfortunately, the architecture nomenclature for LLMs is somewhat confusing, for the purpose of clarification, the encoder-only architecture is an encoder and a decoder just not an auto-regressive decoder and the decoder architecture is an auto-regressive encoder-decoder and the encoder-decoder network really means an encoder with an auto-regressive decoder.
In this work, an encoder architecture is developed and used for multi-variate time series data prediction. \textbf{The encoder architecture is simply a stack of transformer blocks and a feed-forward neural network decoder as seen in Figure \ref{fig t3}.} 

\subsection{Turbofan Engine Degradation Simulation datasets.}
The turbofan engine degradation simulation dataset is one of the datasets on the NASA Prognostics Center of Excellence dataset Repository. This dataset was generated in 2008 in an effort to further research in the field of prognostics and health management by addressing the issue of the lack of common benchmark datasets that researchers can use to compare their approaches.\\
These datasets were generated using the Commercial Modular Aero-Propulsion System Simulation (C-MAPSS). The CMAPPS is a tool hosted on MATLAB and Simulink for simulating realistic large commercial turbofan engines. Four different sets \{FD001, FD002, FD003, FD004\} were simulated under different combinations of operational conditions and fault modes. These datasets are the degradation data of an aircraft turbo engine and were provided as training and test datasets to the competitors in the first prognostics challenge competition at the International Conference on Prognostics and Health Management (PHM08). The challenge remains open for the researchers to develop and compare their efforts.\\
In this work, these datasets are chosen as the use case. Each dataset consists of multiple multi-variate, 26-dimensional time series data from a fleet of engines of the same type, working under similar conditions. There are three operational settings that have a substantial effect on engine performance. Each dataset is further divided into training and test subsets.\\
Each set of data also has distinct characteristics, FD001 is operated in just one condition and has one fault mode, FD002 has six distinct operational settings and one fault mode, and FD003 and FD004 both have two fault modes; one and six operational conditions respectively.\\
Each engine starts with different degrees of initial wear and manufacturing variation which is unknown to the user. This wear and variation is considered normal, i.e., it is not considered a fault condition. The 26-dimensional data consists of the operational settings as the first three dimensions of the dataset, 21 sensor information, and the unit ID and time cycle information are contained in the remaining two dimensions. Table \ref{tab:1} below shows the number of train and test trajectories and the respective operating conditions and fault modes of each set of the C-MAPPS data.\\
\begin{table}[h!]
    % \large
    \centering
    \begin{tabular}{lcccc}
    \hline
     Subsets & No training trajectories & No test trajectories & Operating conditions & Fault modes\\
     \hline
     FD001	& 100 & 100 & 1 & 1 \\
     % \hline
     FD002 & 260 & 259 & 6 & 1\\
     % \hline
     FD003 & 100 & 100 & 1 & 2\\
     % \hline
     FD004 & 249 & 248 & 6 & 2\\
     \hline
    \end{tabular}
    \caption{Description of the C-MAPPS dataset.}
    \label{tab:1}
\end{table}

\subsection{Evaluation Metrics}
Two evaluation metrics are used in this work to evaluate the performance of the encoder transformer model against other state-of-the-art methods. The root mean square error (RMSE) and the scoring metric specifically developed for this dataset in the Prognostics and Health Management (PHM08) competition are used.\\
The RMSE can be expressed as follows:
\begin{equation}
    \sqrt{\sum_{i=1}^{D}(y-y_i)^2}\label{eqn s1}
\end{equation}
where $y_i$ and $y$ are the predicted and true RUL values respectively and N represents the number of test samples.\\
The scoring function proposed in the PHM08 data competition penalizes late predictions more (i.e., the estimated RUL is greater than the true RUL). In the prognostic’s context, the main idea is to avoid failures, it is more desirable to predict early rather than later because if the machine is predicted to fail later than when it fails, costs related to unplanned maintenance actions which are usually greater than planned maintenance costs will be incurred. But if the machine fails later than it was predicted to fail, the facility would already have maintenance plans in place. Therefore, the scoring algorithm for this challenge was asymmetric around the true time of failure such that late predictions were more heavily penalized than early predictions. In either case, the penalty grows exponentially with increasing errors between the true and predicted RULs. The asymmetric preference is controlled by parameters $a_1$ and $a_2$ in the scoring function given in equation \ref{eqn s2} and Figure \ref{fig s1} show the score and the RMSE functions as a function of the error.
\begin{equation}
    s = 
    \begin{cases}
       \sum_{i=1}^{n}e^{-(\frac{d}{a_1})} - 1 & \text{for d $<$ 0}\\\label{eqn s2}
       \sum_{i=1}^{n}e^{(\frac{d}{a_2})} - 1 & \text{for $d \ge 0$}\\
    \end{cases}
\end{equation}
where\\
    {$s$ is the computed score,}\\
    $n$ is the number of machine instances,\\
    $d$ = Estimated RUL - True RUL,\\
    $a_1$ = 10, $a_2$, = 13.\\

\newpage

\section{Proposed Framework and Experimental Study}
\subsection{Application-Specific Details}
\begin{figure}[h!]
    \centering
    \begin{subfigure}[b]{0.45\textwidth}
        \centering
        \includegraphics[width=\textwidth]{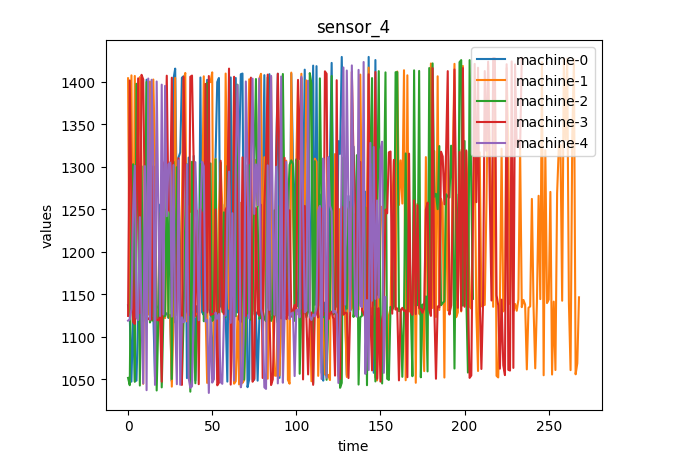}
        \caption{Run-to-failure plots of first 5 ensemble members for sensor 4}
        \label{fig sbd_1}  
    \end{subfigure}
    \hfill
        \begin{subfigure}[b]{0.45\textwidth}
        \centering
        \includegraphics[width=\textwidth]{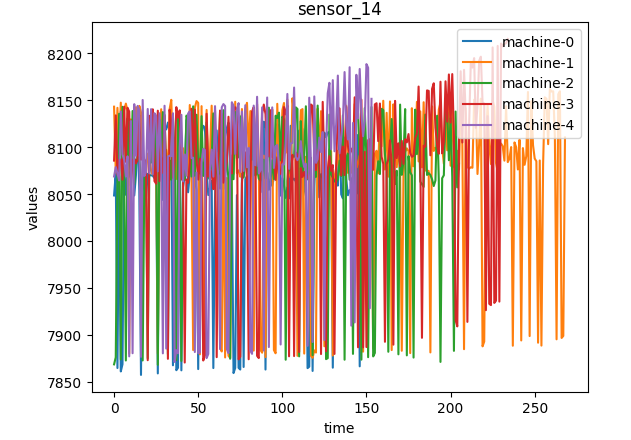}
        \caption{Run-to-failure plots of first 5 ensemble members for sensor 14}
        \label{fig sbd_2}  
    \end{subfigure}
    \caption{Run-to-failure plots before feature extraction.}
    \label{fig sbd}
\end{figure}

\begin{figure}[h!]
    \centering
    \includegraphics[width=0.7\textwidth]{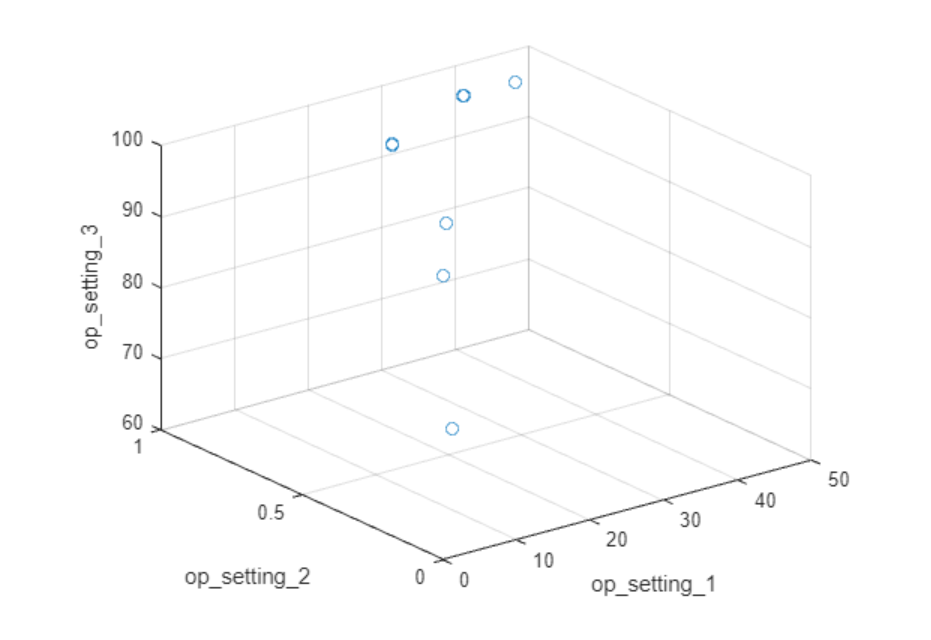}
    \caption{Clusters of operating conditions for datasets FD002 and FD004.}
    \label{fig n1}
\end{figure}

\begin{figure}[h!]
    \centering
    \begin{subfigure}[b]{0.45\textwidth}
        \centering
        \includegraphics[width=\textwidth]{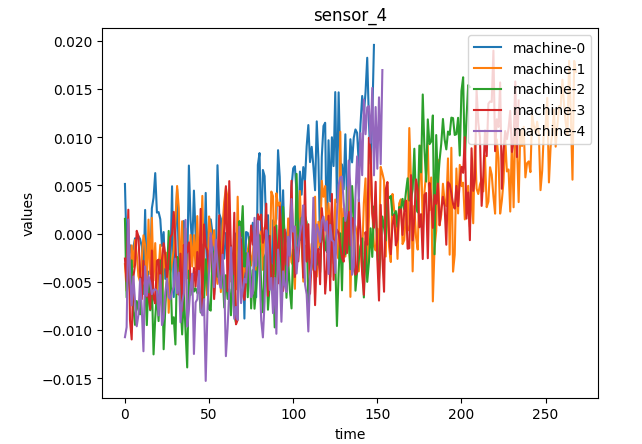}
        \caption{Run-to-failure plots of first 5 ensemble members for sensor 4}
        \label{fig sbdb_1}  
    \end{subfigure}
    \hfill
        \begin{subfigure}[b]{0.45\textwidth}
        \centering
        \includegraphics[width=\textwidth]{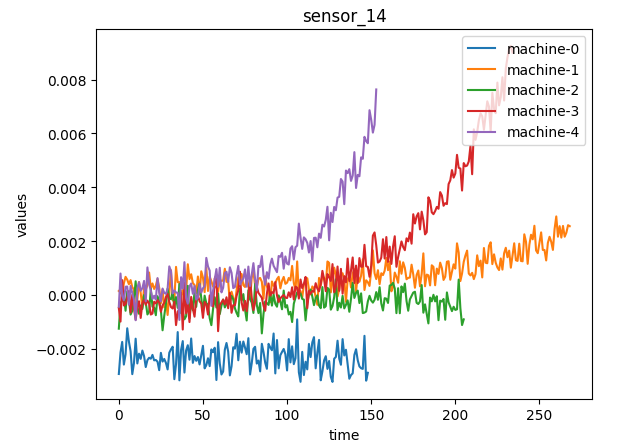}
        \caption{Run-to-failure plots of first 5 ensemble members for sensor 14}
        \label{fig sbdb_2}  
    \end{subfigure}
    \caption{Run-to-failure plots after feature extraction.}
    \label{fig sbdb}
\end{figure}

\begin{figure}[h!]
    \centering
    \includegraphics[width=0.7\textwidth]{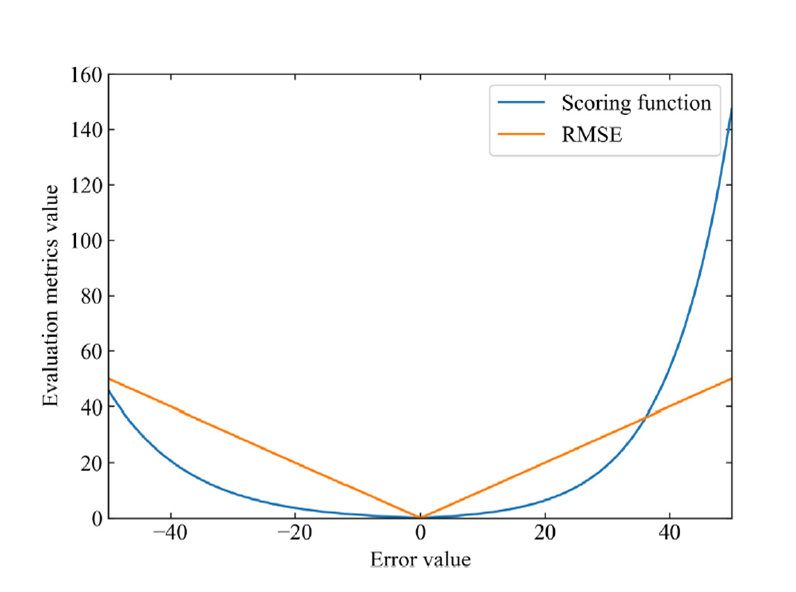}
    \caption{Graphical representation of the scoring function and RMSE. Adapted from \cite{t11}}
    \label{fig s1}
\end{figure}

\textbf{Feature Extraction}\\
Due to the different operational settings and the sensor values calibrations, the degradation trends in the datasets FD002 and FD004 are not obvious. Figure \ref{fig sbd} are the run-to-failure plots from 5 fleets of engines in the FD002 dataset for sensors 4 and 14 chosen at random and there are no clear trends showing the degradation paths of the machines. Clustering and normalization techniques are used to show these trends. 
For every run-to-failure machine instance, the operating conditions are clustered using the k-means clustering algorithm. The greedy k-means algorithm in the sci-kit-learn library was used to find the best number of clusters that resulted in the lowest cost and 6 clusters or modes of operations were found. The multi-dimensional sensor data for each time step are grouped by the clusters their corresponding operational settings belong to, these sensors are then normalized by clusters. The helper function helperPlotClusters in MATLAB was used to visualize the 6 unique clusters of the operational settings on a 3D scatter plot for the FD002 and FD004 datasets with multiple operating conditions. Figure \ref{fig n1} shows the 6 clusters.\\ 
The standardization feature scaling method was used, the mean and standard deviation of each dimension in a cluster are used to normalize the sensor values belonging to that cluster by subtracting the mean and dividing it by the standard deviation. Figure \ref{fig sbdb} are the run-to-failure plots from 5 fleets of engines for sensors 4 and 14 after clustering and normalization, now the degradation trends are more obvious.\\

\textbf{Feature Selection}\\
After the working regimen clustering and normalization processes above, further data analysis was performed to select relevant sensors that give significant degradation information about the machines. Sensors 1,5,6,10,16,18 and 19 showed no trends or had irregular trends so they were not used. Also, the correlation analysis shows that these sensors that show irregular trends are more correlated with the operating conditions than other sensors and they were uncorrelated with other sensors and highly correlated with themselves. More generally, these sensors also have very low correlations to the remaining useful life of the machines, unlike the other sensors that have obvious increasing or decreasing trends from run to failure.\\
Hence in this work, the data from the other 14 sensors, sensors [2,3,4,7,8,9,11,12,13,14,15,17,20,21] were used to train the proposed model(s) and predict the remaining useful life. The sensors used in this work also follow the selected sensors in other studies in the literature [\cite{t14}, \cite{t11}, \cite{t7}].\\

\textbf{Train-Validation Split}\\
The 80-20 train-validation split was used. 80\% of the dataset was used for training and the remaining 20\% for cross-validation. After the hyper-parameters had been chosen, the entire training set was then used to train the final model(s), and the results were evaluated on the test dataset.\\

\textbf{Data Preparation}\\
In this work, data preparation refers to how the data is formatted, shaped, and fed to the model for training and inferencing.\\
The way data is fed into a model can vary from one domain, application, or model to another. This can also impact the performance of the model greatly. For instance, if the data is independently and identically distributed (iid), the input can be prepared in a tabular manner with two dimensions, $N \times D$ where $N$ = Number of data points and $D$ = Number of features. Time series data input $X$ can also be conveniently formatted to have shape $N \times D$. However, this will not usually result in a good performance because one of the ways of enforcing stationarity in time series data is to use a sliding window to wrap portions of the data points for a constant time window. The most preferred format and effect of this way of formatting time series data are further discussed below.\\
As earlier mentioned, because time series data are not iid, they do not have strong-sense stationarity (i.e., the distribution in the stochastic process does not change over time). This affects the way time series data are formatted for use in ML models. To obey the strict assumption that a model can only learn if the data points are iid, non-stationary data are often transformed to become stationary. This is achieved by using a weaker form of stationarity commonly employed in signal processing and time series data analysis known as \textbf{weak-sense stationarity or covariance stationarity}.\\
The weak-sense stationarity implies that the mean and autocovariance don’t change over time. The mean at time $t$ should be the same as the mean at time $t+\tau$ for all $\tau$, and the autocovariance for some random variable $Y$ at time $t_1$ and another random variable $Y$ at time $t_2$ is only a function of the difference between $t_1$ and $t_2$ and only needs to be indexed by one variable, $\tau$ rather than two variables. This equivalently implies that the auto-correlation depends only on $\tau$ = $t_1$ – $t_2$. That is, suppose I take a window of constant length, say $t$ = 20 and I pick any two-time points in the series, if the time difference, $\tau$ is the same, and the covariance is the same, then the data is weak-sense stationary. The relationship between these points in the time series remains constant no matter where I look as long as they are the same distance apart.\\
The weak sense stationarity mean and autocovariance equations can be computed as equations \ref{eqn 8} and \ref{eqn 9} respectively.
\begin{equation}
    \mu_y(t) = \mu_y(t+\tau)  \text{ and for all $\tau$} \label{eqn 8}
\end{equation}
\begin{equation}
    K_{yy}(t_1,t_2) =  K_{yy}(t_1 = t_2, 0)  \text{ and for all $t_1$, $t_2$} \label{eqn 9}
\end{equation}
In time series data analysis, to make the data stationary, the choice of $\tau$ is very important, and it is usually chosen empirically in ML applications. 
In this work, two input data formatting methods were experimented with. 1) Sliding window. 2) Expanding window.\\
\begin{enumerate}
    \item \textbf{Sliding window (Constant Time Window Length):} A sliding window with a constant length and a step size of 1, is used to wrap subsets of the datasets. The time series data input into the transformer is formatted as 3-dimensional data of shape $N \times T \times D$, where $N$ = number of data points, $T$ = sequence length, and $D$ = data dimension or size of the feature space.\\
A detail specific to sequence-to-sequence (seq2seq) modeling is how to deal with varying-length sequences. When datasets of different sequence lengths are considered, the shorter sequences are padded with zeros at the ends to match the length of the longest pre-defined sequence. To ensure that these sequences padded with zeros do not pose as a bottleneck when computing the attention weights, padding masks are used to eliminate and remove their effect.\\
In this data formatting method, since the sequence length $T$ is constant, there was no need to pad the inputs.
Figure \ref{fig t4} shows a schematic of how the sliding window works on time series data.
\begin{figure}[h!]
    \centering
    \includegraphics[width=0.7\textwidth]{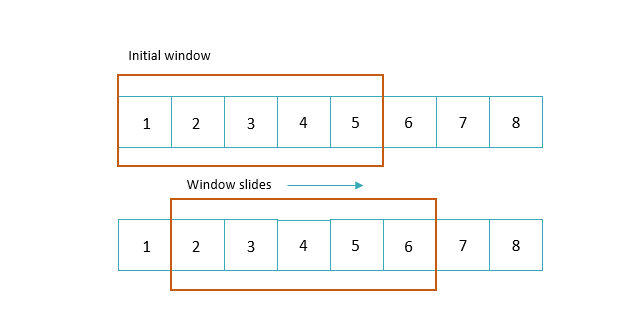}
    \caption{Sliding window process.}
    \label{fig t4}
\end{figure}

    \item \textbf{Expanding window (Increasing time window length):} In this method, a window of increasing length is adopted. An initial minimum window length for every machine instance is set to 5 for the initial data point, then, successive data points are selected by using a window length that is constantly incremented by 1 step size at every time step. Figure \ref{fig t5} shows how the data points are wrapped by the expanding window.\\
    In this case, the sequence lengths are varying so attention padding masks were used.
\begin{figure}[h!]
    \centering
    \includegraphics[width=0.7\textwidth]{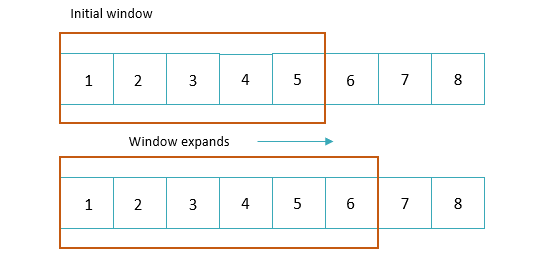}
    \caption{Expanding window process.}
    \label{fig t5}
\end{figure}

\end{enumerate}

\textbf{Experiment Details}\\
The experimentation process can easily become unmanageable due to the huge hyperparameters space, to guarantee a fair comparison between the expanding and sliding window methods, some hyperparameters were set in place. 
The effect of window size of the sliding window method is extremely significant, to cater to this, we used the window sizes used in \cite{t11}. The authors of \cite{t11} already analyzed the impact of the window sizes on the model performance for all four sets of the C-MAPPS dataset.\\
The random and grid search methods were used to find the hyperparameters that resulted in the best model performance. 
While keeping the layer modules constant, the encoder-transformer model hyperparameters that result in the best performances for a given sliding window of length $T$, and the expanding window method for every dataset were found using a random and grid search.
The random search was adopted first to constrain the values of the hyperparameters to a given range then the grid search was used to find the combination of hyperparameters within this range that results in the best model performance.
For the encoder-transformer model, the main hyperparameters that need to be tuned are the model dimensions (\emph{D\_model}), the number of multi-head attention (\emph{n\_heads}), the number of transformer blocks to be stacked upon each other (\emph{n\_transformer blocks}), the dimension of feed-forward network layers (\emph{dim FFW}) and the dropout rate (\emph{dropout rate}). Using the random search, we constrained these values to the range [20-30], [1-4], [1-3], [10-14], and [0.3-0.7] with step sizes 2,1,1,1, and 0.1 respectively. It was observed that the hyperparameters \{D\_model, n\_heads, n\_transformer blocks, dim FFW, dropout rate\} = \{30,2,2,10,0.4\} generally work best for both the sliding and expanding window methods across all the datasets so they were used for the experiments.\\

This also sets the basis for other model-specific experiments conducted in this work. Table \ref{tab:2} shows the hyperparameter used for the experiments.\\
To ensure fairness of the experimentation process, the same training and validation sets were used to perform the experiments for every dataset in the set. 
While keeping other hyperparameters and layers constant, the encoder-transformer model was trained from scratch three times for every experiment.\\
The choice of the evaluation metrics and the dataset upon which the results of the experiments are presented in this work were carefully considered based on the nature of the problem. The RUL score or RMSE values on the test dataset are presented in the experimental result tables because remaining-useful life prediction model(s) generally perform better towards the end-of-life of the machines, this is because the degradation pattern is more obvious at the end of the machine's life, however, at the early stages, the RUL models tend to perform poorly because they have only seen few data points.\\
In the test dataset, we are required to predict the RULs from different stages of the machine's lives; early, mid, or late stages. The test dataset is appropriate because it comes from the same data distribution as the real-world datasets that these models will be applied to and the RUL score function is also a preferred metric for remaining-useful life prediction since early predictions have lower financial and safety implications. \\
All experiments were performed using the NVIDIA Tesla P100 GPU with 16GB memory.

\begin{table}[h!]
    % \large
    \centering
    \begin{tabular}{ccccc}
    \hline
     dim model & n\_heads & n\_transformer blocks & dim FFW & dropout rate\\
     \hline
        30 & 2 & 2 & 10 & 0.4\\
     \hline
    \end{tabular}
    \caption{The encoder-transformer architecture hyperparameters used to conduct experiments.}
    \label{tab:2}
\end{table}

\textbf{Expanding vs Sliding Window Results and Discussion}\\
The expanding window method resulted in large performance improvements of the model compared to the sliding window method of constant length as shown in Table \ref{tab:3} below. An average RUL score reduction of 92.3\% across all the datasets was achieved. Datasets FD002 and FD004 showed the greatest improvement with a percentage increase of 156.6\% and 83.7\% respectively. The effect of the expanding window is less obvious in datasets FD001 and FD003, the reason for this is not known and it remains an area of future work. These results show the effect of this data preparation method on the performance of the transformer model.
\begin{table}[h!]
    % \large
    \centering
    \begin{tabular}{lcccc}
    \hline
      Experiment & FD001 & FD002 & FD003 & FD004\\
     \hline
     Sliding Window	& 661.50	& 1498.82	& 708.21	& 1086.49\\
     % \hline
    \textbf{Expanding Window}	& \textbf{399.50} & \textbf{584.15} & \textbf{436.57}  & \textbf{591.36}\\
     % \hline
     Percentage Improvement (\%)	& 65.58
	& 156.58 & 62.22 &	83.73\\
     \hline
    \end{tabular}
    \caption{Average score values across all datasets for the sliding and expanding window methods of data preparation.}
    \label{tab:3}
\end{table}

This palpable improvement in the performance of the expanding over the sliding window method can be tied to the nature of the degradation paths of machines. One of the most used and well-accepted statistical distributions in reliability engineering, the Weibull distribution introduced in 1939 by Waloddi Weibull divides machine life stages into three, the early failure stage, the constant failure stage, and the wear-out failure stage.  With this understanding, the model’s awareness of the initial stages of the machine’s life will help it to make better predictions and that is what the expanding window method provides. Using a sliding window makes the model lose sight of the initial stages of the machine degradation, it confines the model’s decisions to only t-time steps behind at every point. Also, gradually feeding an incremented sequence length to the model forces it to learn the underlying degradation path of the machines while taking into consideration the initial or early failure stages if there be any for any given machine instance.
It was also observed during training that when using the expanding window method, there was a steady and smoother decrease in the validation loss compared to when using the sliding window method.\\

This method also makes predictions during inferencing more straightforward and built into the model. For the prognostics use case, the test data also has varying sequence lengths, this is because we should be able to determine the remaining useful life of the machine at any time before its failure, be it after 2 days of operation or after 50 days of operation. Using the expanding window length has helped the model to learn to make good predictions with varying length inputs.\\
The input data also has the format $N \times T \times D$ with $T$ increasing from one data point to the other for a given machine instance. To handle the varying length inputs into the transformer model, the inputs with sequence lengths that are less than the maximum sequence length in the training set (the life length of the machine with the longest lifespan in the training dataset) are padded with zeros.\\ 

\begin{figure}
    \centering
    \includegraphics{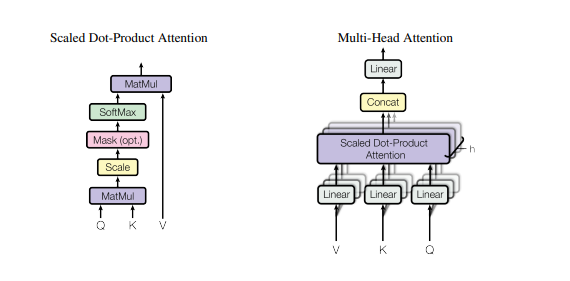}
    \caption{Scaled-Dot Product Attention Mechanism. {Adapted from \cite{t6}}}
    \label{fig t6}
\end{figure}

Finally, attention masks of shape $N \times T$ are also passed as inputs to the encoder transformer model. The attention masks are required to ensure that the self-attention mechanism does not pay attention to the padded tokens/inputs because they do not add any relevant information to the model. This is achieved by creating an $N \times T$ padding array mask for every data point and setting the sequences we want the model to pay attention to one and the padded sequences in the input $X$ to zero.\\
Figure \ref{fig t6} adapted from the paper “Attention is all you need” by \cite{t6}, shows how the self-attention mechanism is computed with an optional choice of attention masks when working with varying-length input sequences.\\

\begin{figure}
    \centering
    \includegraphics[width=0.7\textwidth]{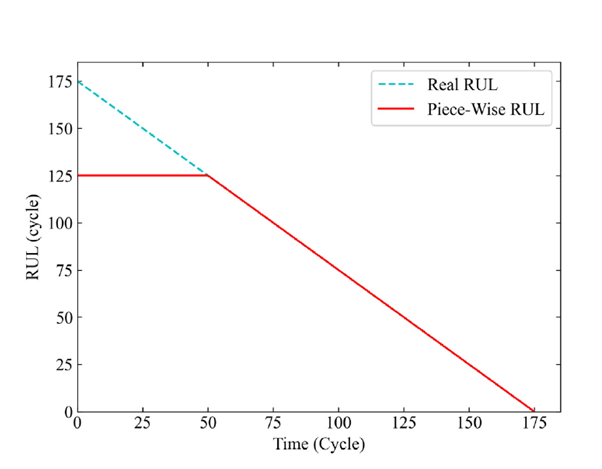}
    \caption{Rectified piecewise remaining useful life function.}
    \label{fig t7}
\end{figure}

\textbf{Target data preparation}\\
The target, $Y$ must also be considered and effectively passed into the model. In this model, the remaining useful life or survival duration from any point in time is the target. General machines don’t just fail abruptly, they usually first operate normally if they are not subjected to early-stage failures, and they deteriorate in some manner. In the following previous works [\cite{t7}, \cite{t8}, \cite{t9}, \cite{t10}, \cite{t11}], a piece-wise RUL was employed rather than the actual RUL as seen in Figure \ref{fig t7}.
Using the degradation profile and the lifetime distributions of all the machines in the datasets, it was observed that the actual RUL label values RUL\_early $\ge $ 125 and remain constant at the beginning and are assumed to degrade linearly until they fail.\\
Following the work in \cite{t11}, to facilitate the training process, we also rescaled the RUL label to the range [0, 1] using the min-max normalization method shown in equation \ref{eqn n1}. Where the \emph{y\_min} is 0 and \emph{y\_max} is 125. During testing, the predicted RUL values are transformed back to values between [125,0].
\begin{equation}
    {y^i}_{new} = \frac{y^i - y_{min}} {y_{max} - y_{min}}
\end{equation}
\subsection{The Encoder-Transformer Model-specific Experiment}\label{sect:4.4}
In the “Attention is all you need” in \cite{t6}, where transformers based solely on self-attention mechanisms were introduced to dispense the need for recurrent and convolution neural networks; the experiments were performed on machine translation tasks. However, in this work, we aim to find ways to efficiently leverage transformer models to make better predictions on time series data. To achieve this and transfer knowledge from the NLP domain, it is expedient that while following the base transformer architecture like using a self-attention mechanism and skip connections, we also need to tailor some of the transformer modules to time series data. In this section, we discuss the model-specific experimentations and hyperparameter tuning process.\\

Three model-specific experiments were conducted, layer normalization versus batch normalization layers, and fixed versus learnable positional encodings. Three input data transformation methods were also tested.
These experiments were carried out using an elimination method i.e. while keeping the other layers of the encoder network and hyperparameters constant, the layers of interest were tuned.
The average RMSE values of the RUL errors on the test datasets were reported for these experiments.

\subsubsection{Layer vs Batch Normalization} For NLP applications as proposed in \cite{t6}, layer normalization is used after the multi-head attention and feed-forward network layers leading to significant performance improvement over batch normalization. In \cite{t2}, it was highlighted that batch normalization resulted in better performance on seven benchmark time series prediction and classification tasks datasets.\\
Table \ref{tab:4} below shows the experimental results of using batch normalization vs layer normalization method and batch normalization resulted in poorer performance on degradation time series data which is contrary to the results observed in the time series classification and regression benchmark datasets used in \cite{t2}.\\
Due to the contradictory results on the different sets of time series data, we cannot conclude that batch normalization is more suitable than layer normalization or vice-versa so it is advised to tune these layers for specific time series applications.
However, for the NASA turbo-engine degradation datasets, we can concretely affirm that layer normalization results in better performance of the transformer model compared to batch normalization with an average improvement increase of over 50\% across all datasets.

\begin{table}[h!]
    % \large
    \centering
    \begin{tabular}{lcccc}
    \hline
      Experiment & FD001 & FD002 & FD003 & FD004\\
     \hline
     Batch normalization	& 22.64	& 20.68	& 23.03	& 25.23\\
     % \hline
    layer normalization	& 17.68 & 12.50 & 18.34  & 12.15\\
     % \hline
     Percentage Improvement (\%)	& 28.01 & 	65.40 &	25.57	& 107.66\\
     \hline
    \end{tabular}
    \caption{The table of results for the layer vs batch normalization layers experiment.}
    \label{tab:4}
\end{table}

\subsubsection{Fixed vs Learnable Positional Encodings} 
The positional encodings can be fixed or learnable parameters. On average as seen in Table \ref{tab:5}, fixed positional encodings result in better results which is consistent with the recommendations in \cite{t2} and \cite{t6}.

\begin{table}[h!]
    % \large
    \centering
    \begin{tabular}{lcccc}
    \hline
      Experiment & FD001 & FD002 & FD003 & FD004\\
     \hline
     Learnable positional encoding	& 17.40	& 13.74	& 25.10	& 13.54\\
     % \hline
    Fixed positional encoding	& 17.68 & 12.50 & 18.34  & 12.15\\
     % \hline
     Percentage Improvement (\%)	& -1.61 &	9.88	& 36.83	& 11.42\\
     \hline
    \end{tabular}
    \caption{The table of results for the fixed vs learnable positional encoding methods experiment.}
    \label{tab:5}
\end{table}

\subsubsection{Input Data Transformations}
In NLP, the input embedding layer converts each token into embeddings, commonly referred to as embedding vectors. This type of representation allows similar words to have similar embeddings. It is a technique that allows individual tokens to be represented as real-valued vectors and this technique is implemented in a way like neural networks, and hence the technique is often lumped into the field of deep learning.\\
This approach of representing words as embedding vectors is considered one of the key breakthroughs of deep learning on challenging NLP problems. For time series problems, researchers try to adopt this idea by using different input transformation methods. In this work, experiments that involve using a linear transformation, convolution layers, and no initial transformations on the input data (i.e., $D_{model}$ = $D_{features}$) were carried out. \\

\textbf{Linear Projection}\\
The linear projection method was introduced in \cite{t2}, and the authors reported better performance with this method.\\
Given that each training sample X $\in \mathbb{R}^{T\times d}$ is a multi-variate time series of length $T$ with $d$ features, constituting a sequence of $T$ feature vectors x\_t $\in \mathbb{R}^{d}$: X $\in \mathbb{R}^{T\times d}$ = $[x_1,x_2,...,x_T]$. The feature vectors $x_t$ are linearly projected onto a $d_{model}$ dimensional vector space, where $d_{model}$ represents the dimension of the transformer model rather than using the original input data features $D_{features}$.
% The linear projection equation \ref{eqn n1}.
\begin{equation}
    u_t = W_px_t + b_p \label{eqn n1}
\end{equation}
where $W_p \in \mathbb{R}^{d_{model}\times d}$, $b_p \in \mathbb{R}^{d_{model}}$ are learnable parameters and $u_t \in \mathbb{R}^{d_{model}}, t = 0,...,T$ are the model input vectors.\\

\begin{figure}
    \centering
    \includegraphics[width=0.7\textwidth]{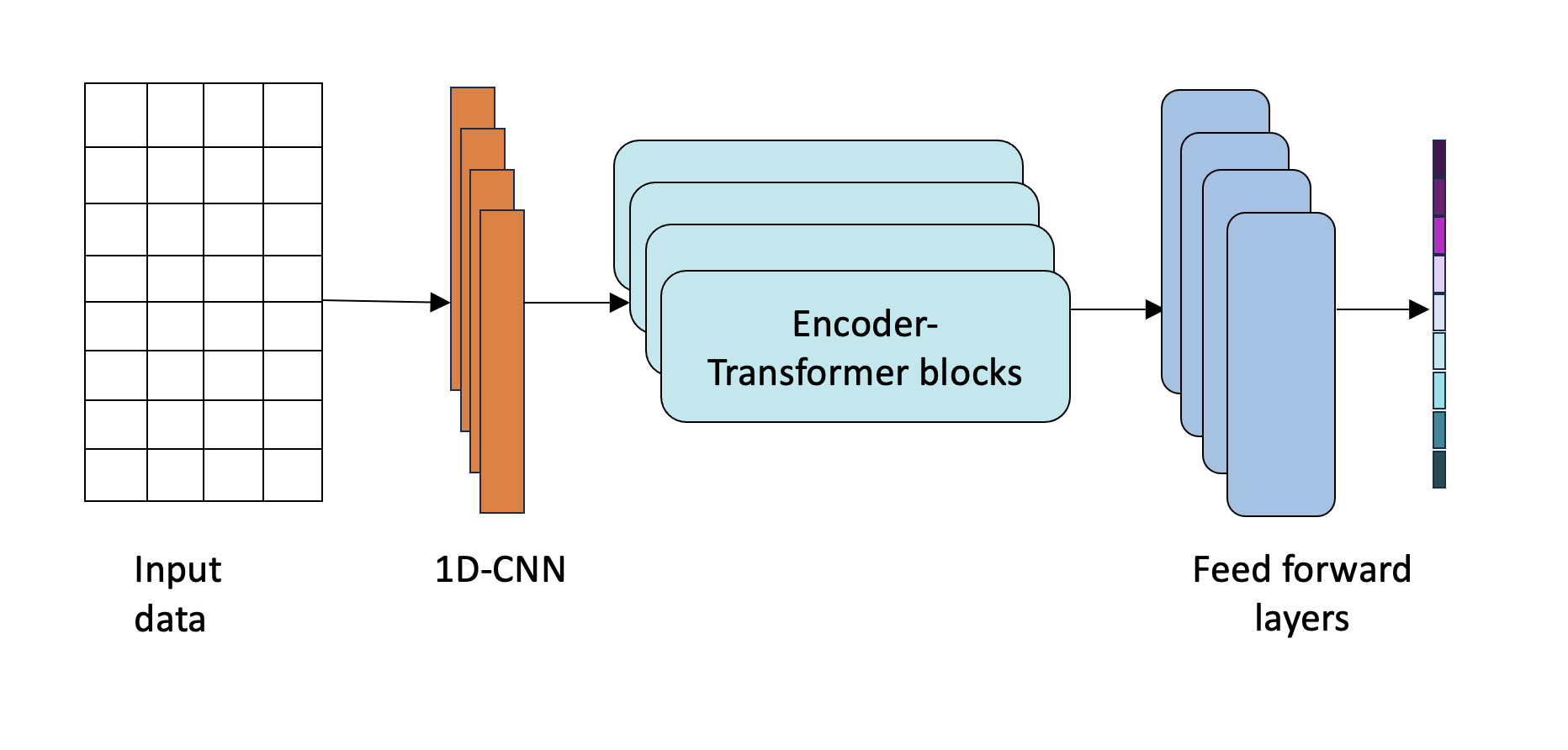}
    \caption{1-CNN with encoder-transformer architecture.}
    \label{fig n2}
\end{figure}

\textbf{1-D CNN for Input Transformation}\\
For this experiment, a single 1-D convolution layer was used over the $TxD$ dimensions for a single training sample. The convolutional kernels are convolved over the temporal dimension for every feature vector $x_d$ of sequence length $T$. This way, the temporal resolution can be controlled and $d_{model}$ will have the size of the number of convolution kernels used. Figure \ref{fig n2} shows the 1D-CNN with the encoder-transformer network.\\

\begin{table}[h!]
    % \large
    \centering
    \begin{tabular}{lcccc}
    \hline
     Experiment & FD001 & FD002 & FD003 & FD004\\
     \hline
     No transformation	& 15.19	 & 26.00	& 24.4 & 31.66\\
     % \hline
     1D-CNN &	22.42 &	26.83 &	42.12	& 28.04\\
     % \hline
     \textbf{Linear Transformation}	& \textbf{17.68} & \textbf{12.50} & \textbf{18.34} & \textbf{12.15}\\
     \hline
    \end{tabular}
    \caption{The table of results using different input data transformation methods.}
    \label{tab:6}
\end{table}

From Table \ref{tab:6}, we can observe that using the linear projection triggered the best model performance. The greatest performance improvements were observed in sets FD002 and FD004 with an average performance increase of 134.3\% over using no input data transformation methods.\\
From the performance improvement rates for FD001 and FD003 with an average of 9.5\%, it can be noted that the input vector $u_t$ need not necessarily be obtained from the linearly transformed feature vectors because the model has quadratic computational complexity. For a better trade-off between model performance and computational efficiency, the input data transformation can be neglected, and the $d_{features}$ used directly for FD001 and FD003 datasets.\\

We observed poorer performance when using a 1D-convolutional layer, this might however be attributed to using a single 1D-convolution layer with 6 kernels for the input data transformation. Perhaps, using deeper convolution layers for the input data transformation might result in a better performance. This is not explored deeper in this work because the objective of this work is to show the performance of the native encoder-transformer architectures without deep recurrent or convolution layers.\\
It was observed that using the data directly without transformations resulted in relatively high RMSE values and low score values compared to the 1-D CNN transform method with very high score values. And fewer improvements were observed between using the linear transformation method vs no transformation method for the FD001 and FD003 datasets.\\
This result is intuitively reasonable because of the self-attention mechanisms used in the encoder-transformer network. The self-attention mechanism is essentially a combination of dot products on sequences from the same data. The premise is, the dot product is a measure of similarity, the values of the dot product are large when two vectors are similar and low when the vectors are dissimilar. In time series data prediction, we leverage the auto-correlation between the features to make predictions and the dot product is analogous to correlation, so using the multi-head self-attention mechanism directly on the input data can find the auto-correlations in the multi-dimensional vector space resulting in the model learning well from the data directly.\\

\subsection{Proposed RUL Prediction Method}
Figure \ref{fig n3} below shows the final workflow of the proposed RUL prediction method for the NASA turbo-engine datasets.
\begin{figure}[h!]
    \centering
    \includegraphics[width=\textwidth]{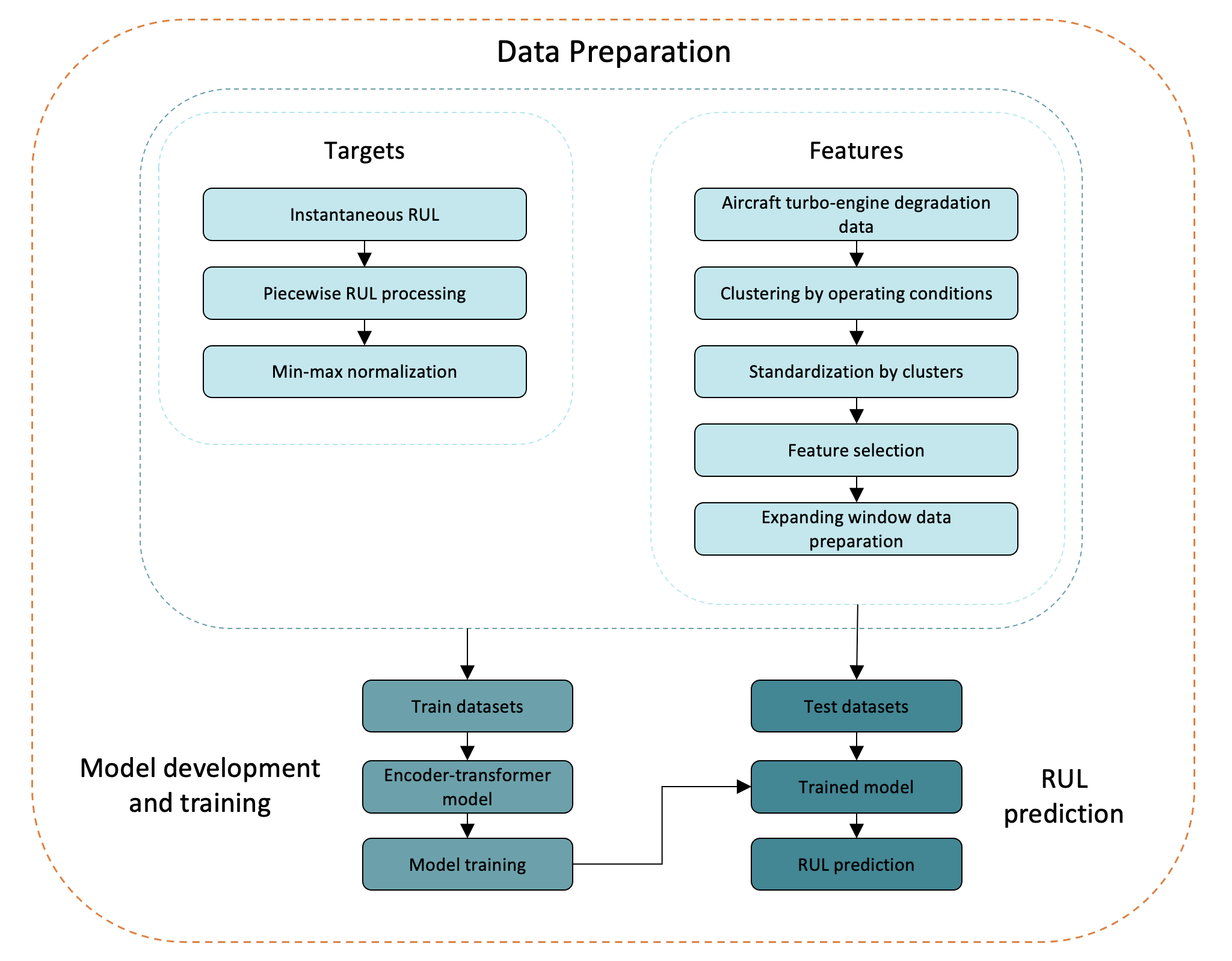}
    \caption{Workflow of the proposed RUL prediction method.}
    \label{fig n3}
\end{figure}
% \subsection{Hyperparameter Selection}
The layer normalization, fixed positional encoding, and linear transformation were used in the final encoder-transformer model. The hyperparameters of the models were then tuned for the specific datasets and used to train the final encoder-transformer models. Table \ref{tab:7} shows the best hyperparameters for each dataset.
\begin{table}[h!]
    % \large
    \centering
    \begin{tabular}{lcccc}
    \hline
     Hyperparameters & FD001 & FD002 & FD003 & FD004\\
     \hline
     dim model	& 18 & 26 & 22 & 26 \\
     % \hline
     n\_heads	& 2 & 2 & 2 & 2\\
     % \hline
     n\_transformer blocks	& 1 & 2 & 2 & 1\\
     % \hline
     dim FFW & 8 & 10 & 10 & 10 \\
     % \hline
     dropout rate	& 0.4 & 0.4 & 0.4 & 0.4\\
     \hline
    \end{tabular}
    \caption{Best hyperparameters for the encoder-transformer architecture.}
    \label{tab:7}
\end{table}

\newpage
\section{Evaluation, Analysis, and Comparisons}
\label{chapter:eval}
\subsection{Encoder-transformer-based Model Results} \label{sect:5.2}
\begin{figure}[h!]
    \centering
    \begin{subfigure}[b]{0.45\textwidth}
        \centering
        \includegraphics[width=\textwidth]{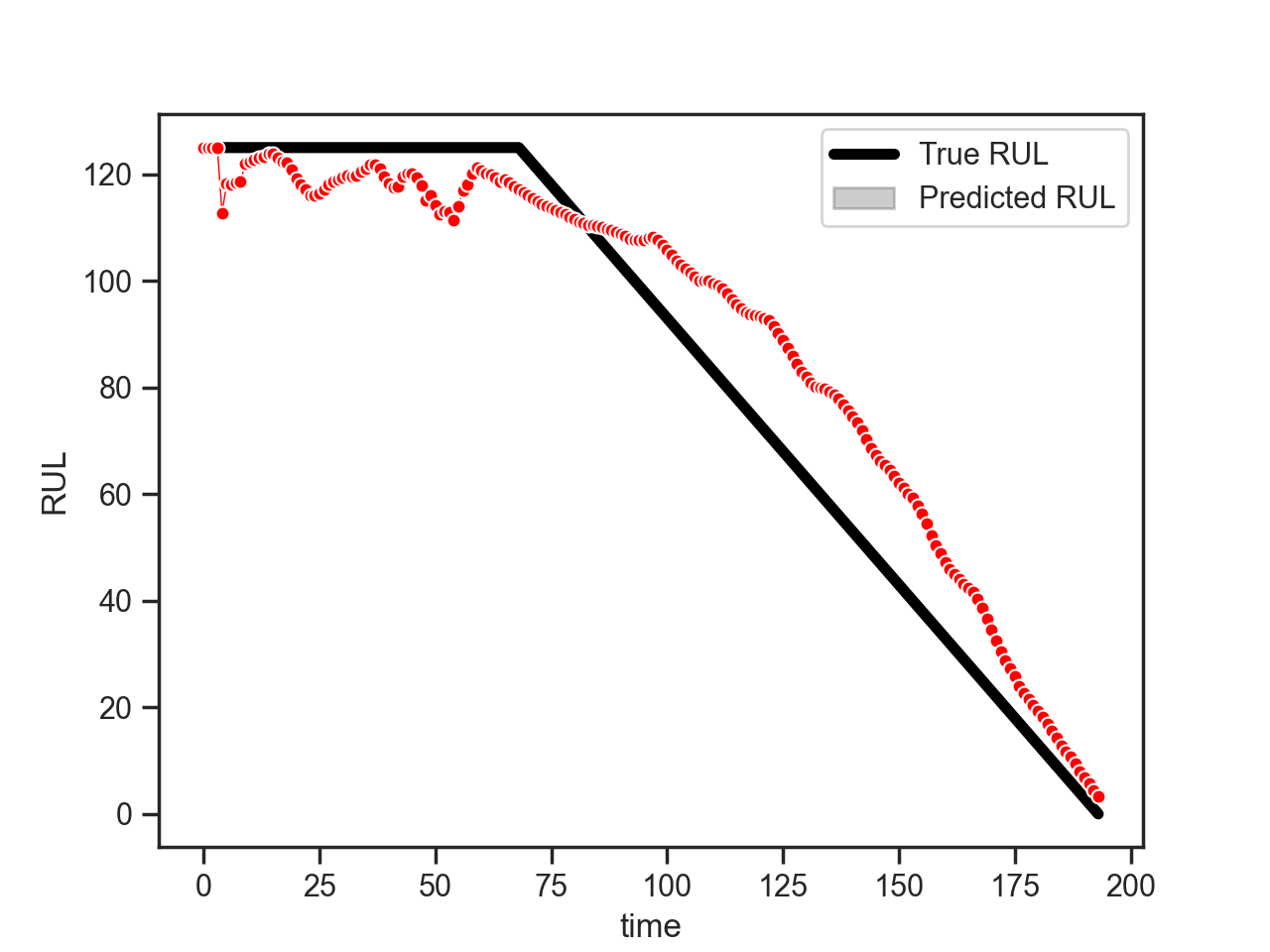}
        \caption{Prediction result of engine 205}
        \label{fig t9a}  
    \end{subfigure}
    \hfill
        \begin{subfigure}[b]{0.45\textwidth}
        \centering
        \includegraphics[width=\textwidth]{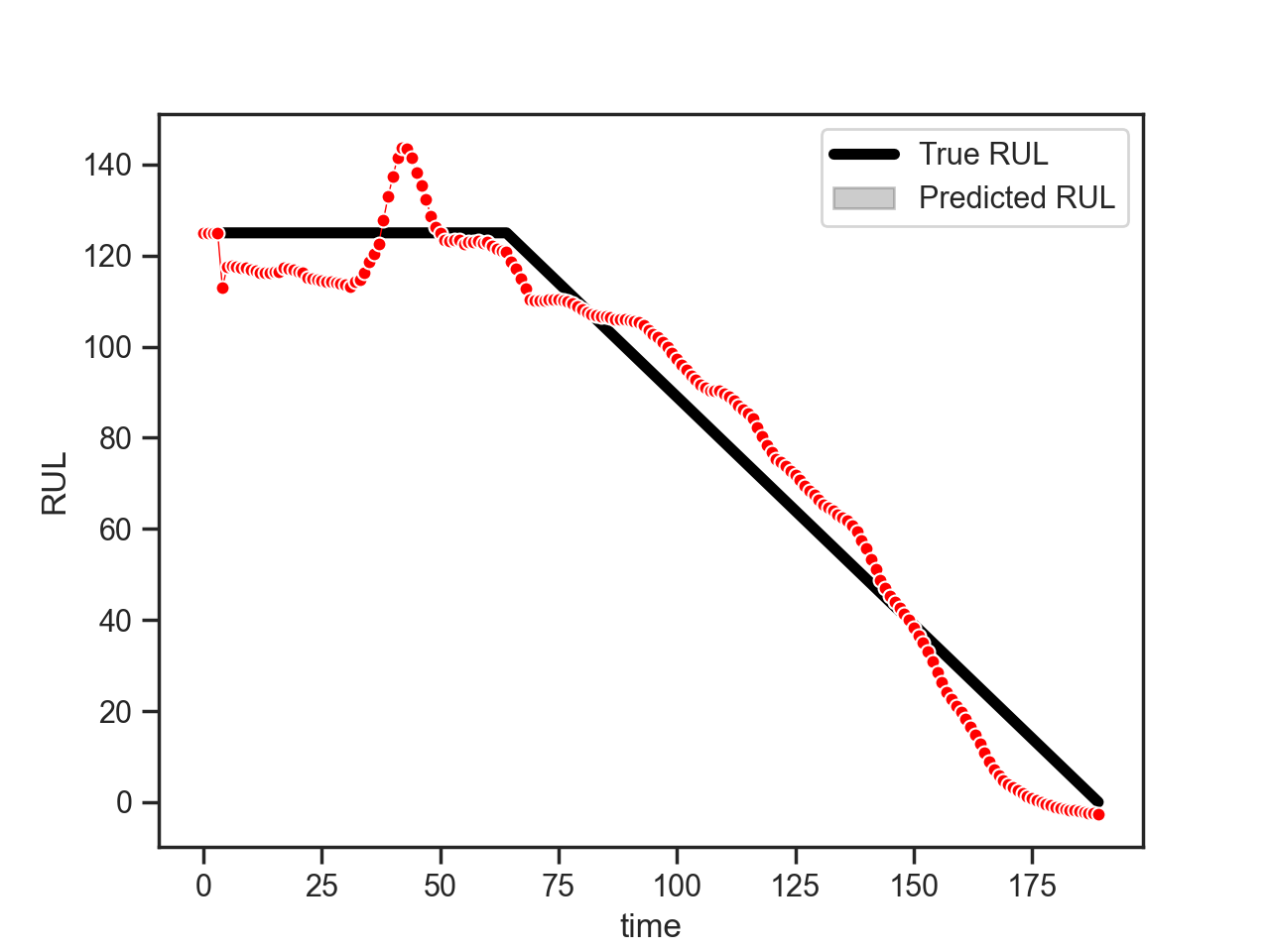}
        \caption{Prediction result of engine 246}
        \label{fig t9b}  
    \end{subfigure}
    \caption{Predicted and true RUL plots for two engines in FD002 validation dataset.}
    \label{fig 9}
\end{figure}

\textbf{Visualization of the predicted vs true RULs}\\
To evaluate this model, the RUL predictions on a set of engine units picked randomly from the FD002 validation dataset are plotted in Figure \ref{fig 9}. The estimated RULs have the same downward trend as the true RULs after RUL\_early till failure and the predicted RUL tracks the true RUL very closely at every time step. Also, the predicted RULs at the early life are near to the constant RUL\_early at the beginning of the unit life.\\

\begin{figure}[h!]
    \centering
    \includegraphics[width=0.7\textwidth]{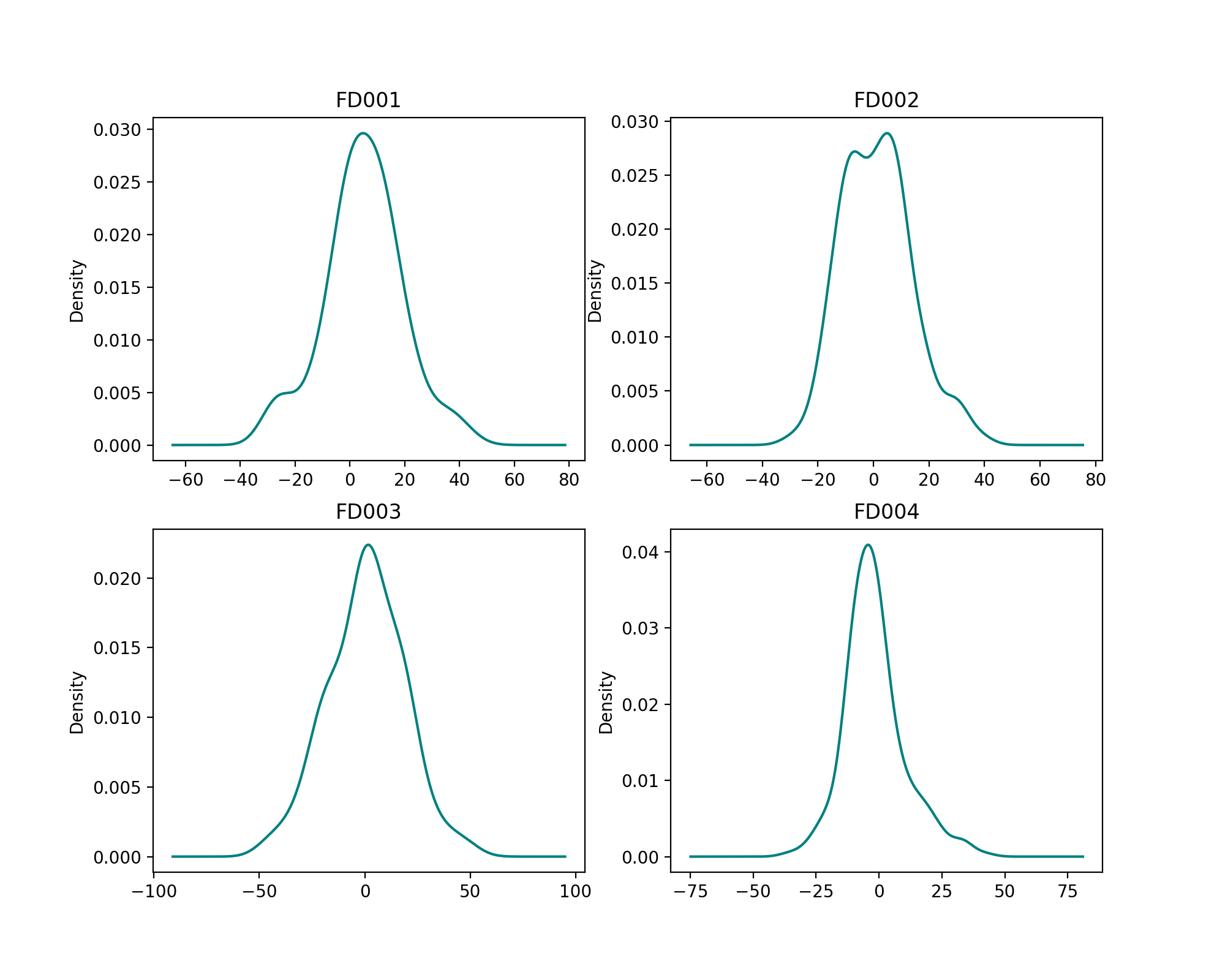}
    \caption{Probability distribution plots of the RUL errors.}
    \label{fig 12}
\end{figure}

\begin{figure}[h!]
    \centering
    \includegraphics[width=0.7\textwidth]{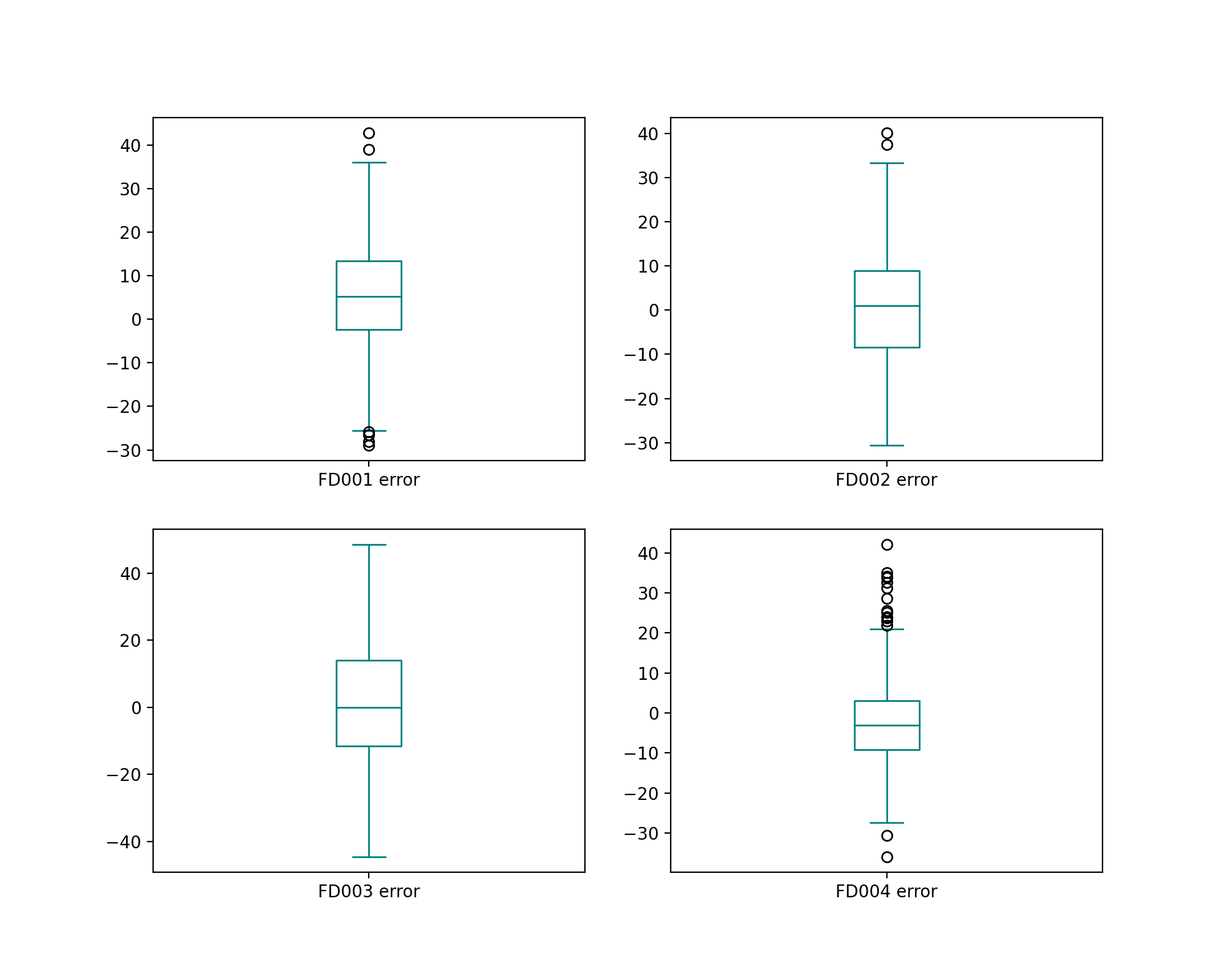}
    \caption{Box plots of the RUL errors.}
    \label{fig n12}
\end{figure}

\textbf{True vs Predicted RUL Error Statistical Analysis and Results Interpretation.}\\
Descriptive statistics of the RUL errors $d$ (d = predicted RULs - true RULs ) were conducted. Probability distribution and box plots of the errors between the true and the predicted RULs were plotted in Figures \ref{fig 12} and \ref{fig n12} respectively.\\

Table \ref{tab:8} below shows the mean, mode, median, skewness coefficient, and standard deviation of the RUL errors.\\
From the probability distribution plots, the errors for FD001 and FD003 datasets are relatively symmetric around the mean and slightly left skewed. The FD002 and FD004 datasets errors are moderately right skewed which means there are probably more outliers on the upper bound of the data distribution. It shows that most predictions are early prognoses and the outliers come from late RUL predictions. The box plots clearly show the outliers.\\
The intensity of skewness in Table \ref{tab:8} shows how much deviation the probability distributions are from a normal probability distribution and if the error distribution can be adequately represented with the mean and standard deviation. The Pearson coefficient of skewness was used to check the degree of skewness and the coefficient of skew for all the datasets were between  -0.5 and 0.74, so the errors are fairly symmetrical and not badly skewed.\\ \textbf{Because the RUL errors are moderately skewed, the error population can be represented with the median and standard deviation of the RUL errors.}
\begin{table}[h!]
    \centering
    \begin{tabular}{lcccc}
    \hline
      Statistics &  FD001 & FD002 & FD003 & FD004\\
     \hline
     Mean & 5.05 & 1.28 & 0.72 & -1.72\\
     % \hline
     Median & 5.24 & 1.00 & 0.0 & -3.07\\
     % \hline
     Std & 14.20 & 12.71 & 18.01 & 11.99\\
     % \hline
     Skew coefficient & -0.06 & 0.42 & -0.06 & 0.74\\
     \hline
    \end{tabular}
    \caption{Descriptive statistics table for RUL errors.}
    \label{tab:8}
\end{table}

\subsection{Discussion of Results} 
\begin{figure}[h!]
    \centering
    \begin{subfigure}[b]{0.45\textwidth}
        \centering
        \includegraphics[width=\textwidth]{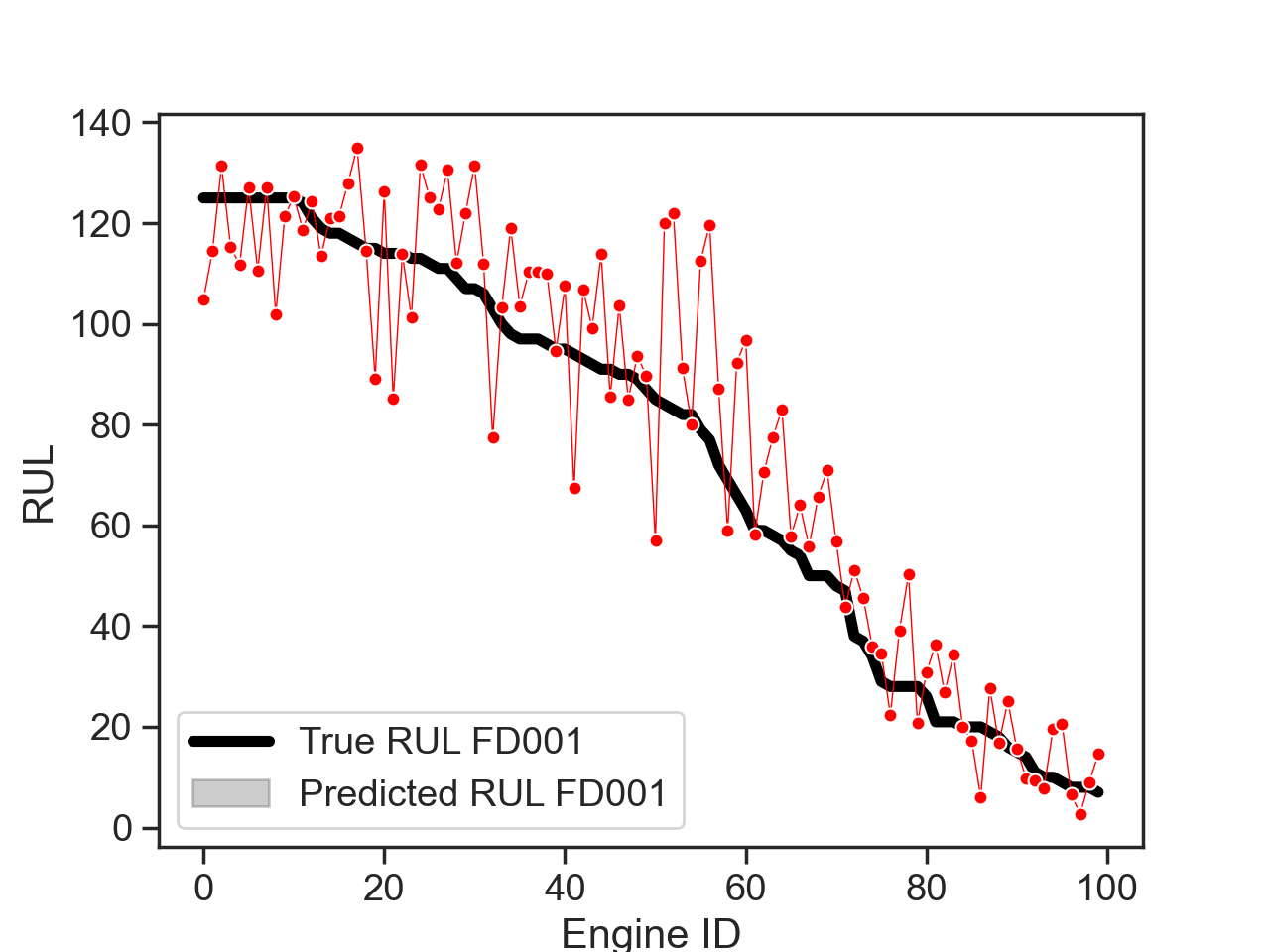}
        \caption{True RUL vs predicted RUL on FD001 dataset.}
        \label{fig nt12a}  
    \end{subfigure}
    \hfill
        \begin{subfigure}[b]{0.45\textwidth}
        \centering
        \includegraphics[width=\textwidth]{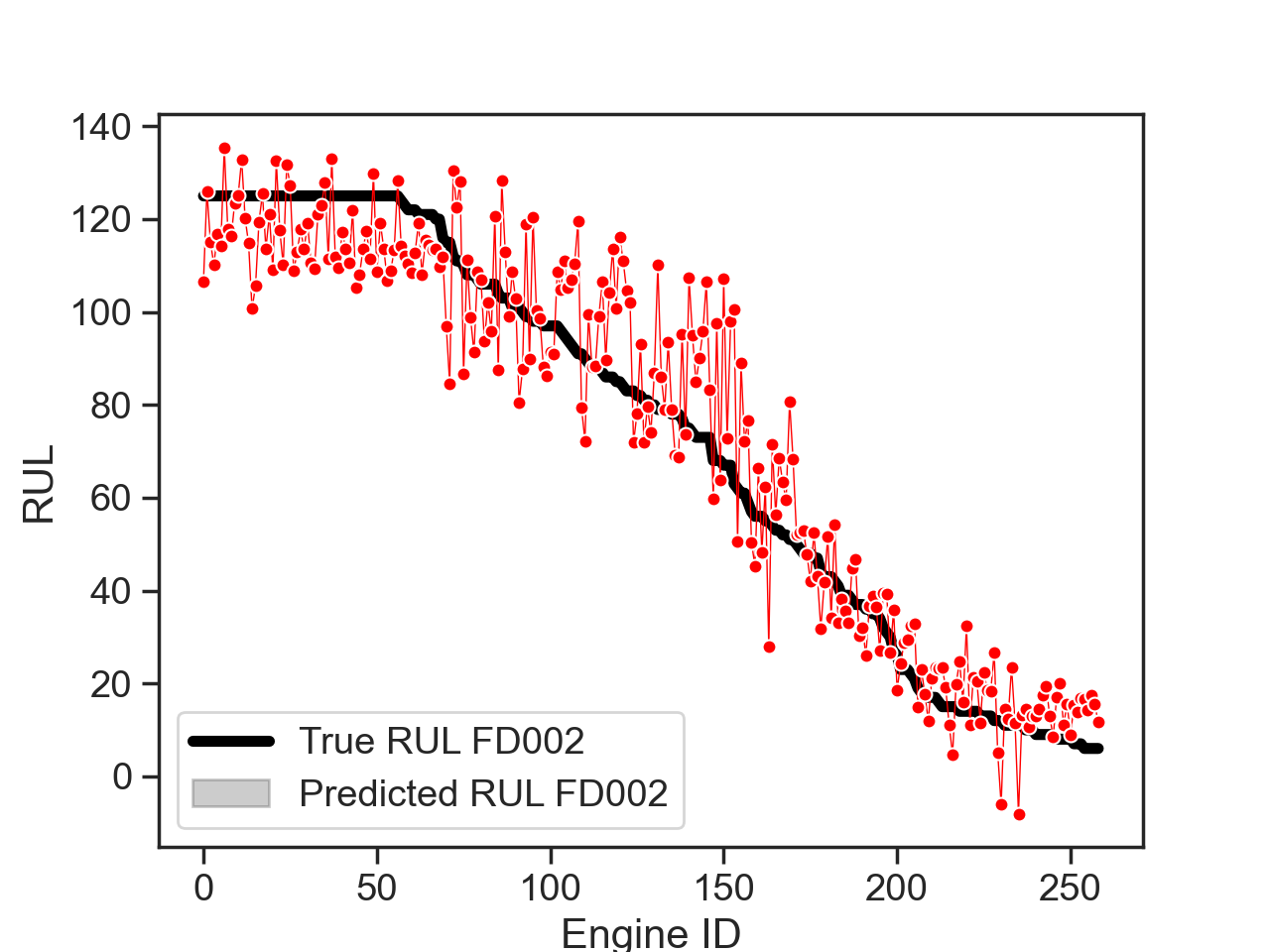}
        \caption{True RUL vs predicted RUL on FD002 dataset.} 
        \label{fig nt12b}
    \end{subfigure} 
    \hfill
        \begin{subfigure}[b]{0.45\textwidth}
        \centering
        \includegraphics[width=\textwidth]{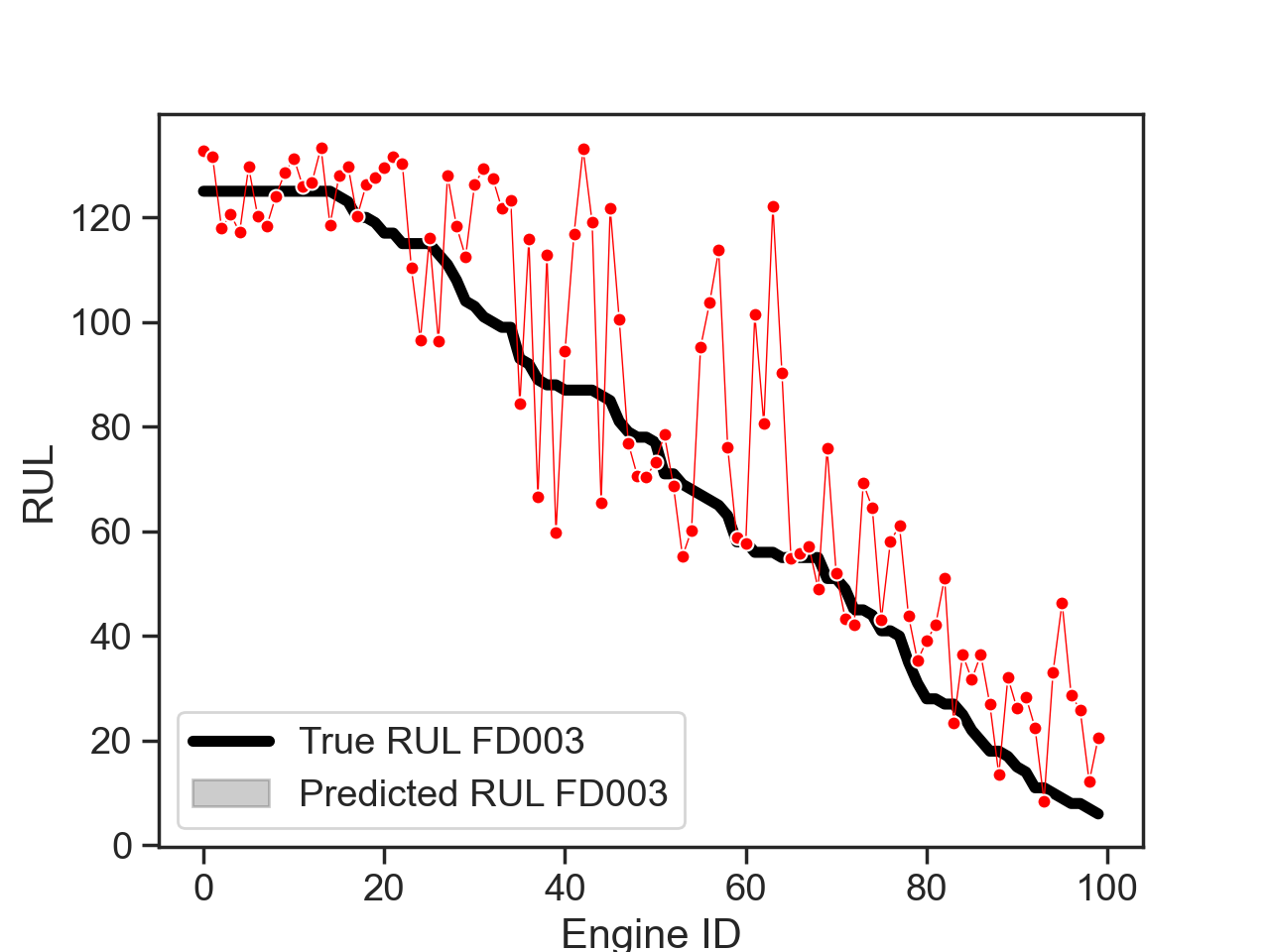}
        \caption{True RUL vs predicted RUL on FD003 dataset.}
        \label{fig nt12c}  
    \end{subfigure}
        \hfill
        \begin{subfigure}[b]{0.45\textwidth}
        \centering
        \includegraphics[width=\textwidth]{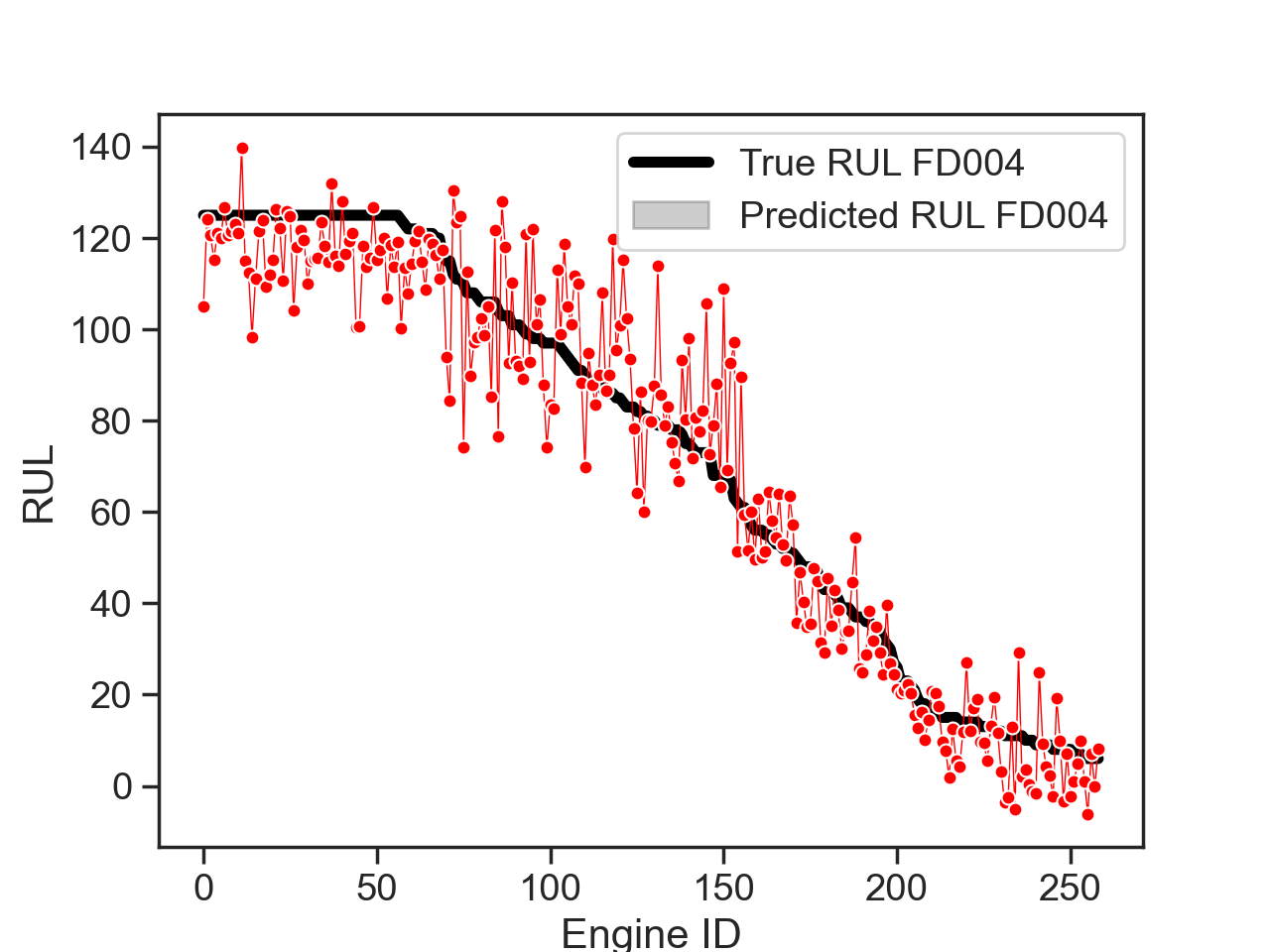}
        \caption{True RUL vs predicted RUL on FD004 dataset.}
        \label{fig nt12d}  
    \end{subfigure}
    \caption{True RUL vs predicted RUL on the test datasets.}
    \label{fig t11}
\end{figure}

Finally, the performance of the proposed encoder-transformer model was evaluated on the test dataset with the RMSE, and score function metrics; and compared with the results from other state-of-the-art models in the literature. The proposed encoder-transformer model surpassed the 13 other models it was compared with on the FD002 and FD004 datasets.\\
An average performance increase of 90.5\% on the score function was achieved over the current best scores in the literature \cite{t7} on the FD002 and FD004 datasets. These results show the effectiveness of the native encoder-transformer model without recurrent or convolution networks. It shows a huge improvement in performance compared to the other augmented transformer models that have been developed for this dataset. These improvements are also a result of the expanding window method proposed in this work, the impact of the expanding window method is not prominent on the FD001 and FD003 datasets, this might also be the reason why exceeding improvements in performance are not observed on them.\\
It shows that having a larger model does not necessarily guarantee a better result on the unseen dataset. However, these results are also a precursor to future improvements that can be made to this model, an augmented transformer model with the expanding window method proposed in this work can be used in the future with the current setup, and better results are hypothetically expected.\\
The true versus predicted RUL values on the test dataset are plotted in Figure \ref{fig t11}. As seen on these plots, the predicted RULs track the true RULs very closely across all datasets, especially in FD002 and FD004, which is why their RMSE RUL error values are very low. More interestingly, in the FD002 plot, more early predictions were made so the scores are penalized less giving rise to the very low score values. \\
In FD001 and FD003, the predicted RUL tracks the true RUL more loosely compared with FD002 and FD004, and more late predictions were made resulting in the relatively higher score and RMSE values.

\begin{table}[h!]
    % \large
    \centering
    \begin{tabular}{llllll}
    \hline
      Models & FD001  &  FD002 & FD003 & FD004 & Average\\
       \hline
        HMC \cite{t12}	& 427 & 19 400	& 2977 & 10 376 & 14 888\\
        CNN \cite{t13}	& 1287 & 13 570	& 1596 & 7886 & 6084.75\\
        DCNN \cite{t14} & 274	& 10 412 & 284 & 12 466 & 5859\\
        MODBNE \cite{t15} & 334	& 5585 & 422 & 6558 & 3224.75\\
        LSTM-FNN \cite{t16} & 338	& 4450 & 852 & 5550 & 5000\\
        CatBoost \cite{t17}	& 398.7	& 3493.2 & 584.2 & 3203.4 & 1919.88\\
        RBM-LSTM-FNN \cite{t18}	& 231 & 3366 & \textbf{251} & 2840 & 3103\\
        DFC-LSTM \cite{t19} & N/A & 3296.3	& N/A & N/A & 3296.3\\
        Auto-encoder \cite{t20} & 228 & 2650	& 1727 & 2901 & 3150.5\\
      PF \cite{t21} & 383.39 & 1226.97	& 375.29 & 2071.51 & 1014.29\\
      BI-GRU-TSAM \cite{t23} & N/A & 2264 & N/A & 3610 & 2937\\
     Double-Attention-based architecture	\cite{t11} & \textbf{198} & 1575	& 290 & 1741 & 1661\\
     MSTformer \cite{t7} & N/A & 1099 & N/A & 1012 & 1055.5\\
    \textbf{Encoder-transformer (this work)} & 266.52 &	\textbf{552.74}	& 401.66	& \textbf{555.64} & \textbf{444.14}\\
    \hline
    \end{tabular}
    \caption{Scores of the proposed method and SOTA methods on the test datasets.}
    \label{tab:9}
\end{table}

\begin{table}[h!]
    % \large
    \centering
    \begin{tabular}{llllll}
    \hline
      Models & FD001  &  FD002 & FD003 & FD004 & Average\\
       \hline
        HMC \cite{t12}	& 13.84 & 20.74 & 14.41 & 22.73 & 17.93\\
        CNN \cite{t13}	& 18.45 & 30.29 & 19.82 & 29.16 & 24.43\\
        DCNN \cite{t14} & 12.61 & 22.36 & 12.64 & 23.31 & 17.74\\
        MODBNE \cite{t15} & 15.04 & 25.05 & 12.51 & 28.66 & 20.32\\
        LSTM-FNN \cite{t16} & 16.14 & 24.49 & 16.18 & 28.17 & 21.25\\
        CatBoost \cite{t17}	& 15.8 & 21.4 & 16.0 & 22.4 & 18.90\\
        RBM-LSTM-FNN \cite{t18}	& 12.56 & 22.73 & 12.10 & 22.66 & 17.51\\
        DFC-LSTM \cite{t19} & N/A & 20.30 & N/A & N/A & 20.30\\
        Auto-encoder \cite{t20} & 13.58 & 19.59 & 19.16 & 22.15 & 18.62\\
      PF \cite{t21} & 15.94 & 17.15 & 16.17 & 20.72 & 17.50\\
      BI-GRU-TSAM \cite{t23} & \textbf{11.27} & 18.94 & \textbf{11.42} & 20.47 & 15.53\\
     Double-Attention-based architecture	\cite{t11} & 12.25 & 17.08 & 13.39 & 19.86 & 15.65\\
     MSTformer \cite{t7} & N/A & 14.48 & N/A & 15.03 & 14.755\\
    \textbf{Encoder-transformer (this work)} & 14.21 & \textbf{12.75}	& 15.57	& \textbf{12.09} & \textbf{13.66}\\
    \hline
    \end{tabular}
    \caption{RMSE values of the proposed method and SOTA methods on the test datasets.}
    \label{tab:10}
\end{table}

\newpage

\section{Conclusion and Areas of future work.}
In this work, we propose a novel end-to-end RUL prediction method based on the native encoder-transformer model architecture and an expanding window method of data preparation. We also performed one application-based and three model-specific experiments to adequately adapt the encoder-transformer architecture from Natural Language Processing domain for more effective time series prediction.\\
The proposed method in this work was able to achieve an average increase in performance of 137.65\% over the second-best model in Table \ref{tab:9} with an average remaining useful life prediction score value of 444.14 over all the datasets and an average RMSE score of 13.66.\\
This work shows the prediction capability of the native encoder-transformer architecture with augmentation with recurrent, convolution, or channel-attention layers. The palpable improvement in the proposed model can also be attributed to the expanding window method of data preparation first introduced in this work.\\
This work does not invalidate the use of augmented or improved transformer architectures for time series data prediction as used in previous works, it instead shows the efficacy of the native transformer model for time series data and we propose that augmented transformer architectures should be used in future works with the expanding window method and following the set-up meticulously chosen in this work through experimentation.\\
The expanding window method did not result in a large improvement in the performance of the proposed encoder-transformer model for the FD001 and FD003 datasets, the reason for this behavior is an area worth investigating in future works.

\newpage
\section*{Declaration of Conflicting Interest}
The authors declare that they have no known conflicting financial and personal relationships that have influenced the work reported in this paper.

\section*{Acknowledgement}
We would like to acknowledge the financial support of NTWIST Inc. and Natural Sciences and Engineering Research Council (NSERC) Canada under the Alliance Grant ALLRP 555220 – 20, and research collaboration of NTWIST Inc. from Canada, Fraunhofer IEM, D\"{u}spohl Gmbh, and Encoway Gmbh from Germany in this research.
% \section{Bibliography}

\printcredits
\newpage
%\bibliographystyle{ieeetr}
\bibliographystyle{ieeetr}

\bibliography{cas-refs}

\begin{thebibliography}{10}

\bibitem{t2}
G.~Zerveas, S.~Jayaraman, D.~Patel, A.~Bhamidipaty, and C.~Eickhoff, ``A
  transformer-based framework for multivariate time series representation
  learning,'' in {\em Proceedings of the 27th ACM SIGKDD Conference on
  Knowledge Discovery \& Data Mining}, pp.~2114--2124, 2021.

\bibitem{t1}
A.~Shifaz, C.~Pelletier, F.~Petitjean, and G.~I. Webb, ``Ts-chief: a scalable
  and accurate forest algorithm for time series classification,'' {\em Data
  Mining and Knowledge Discovery}, vol.~34, no.~3, pp.~742--775, 2020.

\bibitem{t3}
J.~Lines, S.~Taylor, and A.~Bagnall, ``Time series classification with
  hive-cote: The hierarchical vote collective of transformation-based
  ensembles,'' {\em ACM transactions on knowledge discovery from data},
  vol.~12, no.~5, 2018.

\bibitem{t4}
A.~Dempster, F.~Petitjean, and G.~I. Webb, ``Rocket: exceptionally fast and
  accurate time series classification using random convolutional kernels,''
  {\em Data Mining and Knowledge Discovery}, vol.~34, no.~5, pp.~1454--1495,
  2020.

\bibitem{r1}
W.~Wang, ``An overview of the recent advances in delay-time-based maintenance
  modelling,'' {\em Reliability Engineering \& System Safety}, vol.~106,
  pp.~165--178, 2012.

\bibitem{ogunfowora2023reinforcement}
O.~Ogunfowora and H.~Najjaran, ``Reinforcement and deep reinforcement
  learning-based solutions for machine maintenance planning, scheduling
  policies, and optimization,'' {\em Journal of Manufacturing Systems},
  vol.~70, pp.~244--263, 2023.

\bibitem{r4}
H.~Wang, Q.~Yan, and S.~Zhang, ``Integrated scheduling and flexible maintenance
  in deteriorating multi-state single machine system using a reinforcement
  learning approach,'' {\em Advanced Engineering Informatics}, vol.~49,
  p.~101339, 2021.

\bibitem{t6}
A.~Vaswani, N.~Shazeer, N.~Parmar, J.~Uszkoreit, L.~Jones, A.~N. Gomez,
  {\L}.~Kaiser, and I.~Polosukhin, ``Attention is all you need,'' {\em Advances
  in neural information processing systems}, vol.~30, 2017.

\bibitem{t22}
J.~Zhang, Y.~Jiang, S.~Wu, X.~Li, H.~Luo, and S.~Yin, ``Prediction of remaining
  useful life based on bidirectional gated recurrent unit with temporal
  self-attention mechanism,'' {\em Reliability Engineering \& System Safety},
  vol.~221, p.~108297, 2022.

\bibitem{t11}
L.~Liu, X.~Song, and Z.~Zhou, ``Aircraft engine remaining useful life
  estimation via a double attention-based data-driven architecture,'' {\em
  Reliability Engineering \& System Safety}, vol.~221, p.~108330, 2022.

\bibitem{sb1}
T.~Wang, J.~Yu, D.~Siegel, and J.~Lee, ``A similarity-based prognostics
  approach for remaining useful life estimation of engineered systems,'' in
  {\em 2008 international conference on prognostics and health management},
  pp.~1--6, IEEE, 2008.

\bibitem{t13}
G.~Sateesh~Babu, P.~Zhao, and X.-L. Li, ``Deep convolutional neural network
  based regression approach for estimation of remaining useful life,'' in {\em
  Database Systems for Advanced Applications: 21st International Conference,
  DASFAA 2016, Dallas, TX, USA, April 16-19, 2016, Proceedings, Part I 21},
  pp.~214--228, Springer, 2016.

\bibitem{t14}
X.~Li, Q.~Ding, and J.-Q. Sun, ``Remaining useful life estimation in
  prognostics using deep convolution neural networks,'' {\em Reliability
  Engineering \& System Safety}, vol.~172, pp.~1--11, 2018.

\bibitem{t10}
H.~Li, W.~Zhao, Y.~Zhang, and E.~Zio, ``Remaining useful life prediction using
  multi-scale deep convolutional neural network,'' {\em Applied Soft
  Computing}, vol.~89, p.~106113, 2020.

\bibitem{rnn1}
J.~Liu, F.~Lei, C.~Pan, D.~Hu, and H.~Zuo, ``Prediction of remaining useful
  life of multi-stage aero-engine based on clustering and lstm fusion,'' {\em
  Reliability Engineering \& System Safety}, vol.~214, p.~107807, 2021.

\bibitem{rnn2}
Q.~Wu, K.~Ding, and B.~Huang, ``Approach for fault prognosis using recurrent
  neural network,'' {\em Journal of Intelligent Manufacturing}, vol.~31,
  pp.~1621--1633, 2020.

\bibitem{rnn3}
Y.~Wu, M.~Yuan, S.~Dong, L.~Lin, and Y.~Liu, ``Remaining useful life estimation
  of engineered systems using vanilla lstm neural networks,'' {\em
  Neurocomputing}, vol.~275, pp.~167--179, 2018.

\bibitem{rnn4}
J.~Zhang, P.~Wang, R.~Yan, and R.~X. Gao, ``Long short-term memory for machine
  remaining life prediction,'' {\em Journal of manufacturing systems}, vol.~48,
  pp.~78--86, 2018.

\bibitem{t16}
S.~Zheng, K.~Ristovski, A.~Farahat, and C.~Gupta, ``Long short-term memory
  network for remaining useful life estimation,'' in {\em 2017 IEEE
  international conference on prognostics and health management (ICPHM)},
  pp.~88--95, IEEE, 2017.

\bibitem{t19}
C.~Wang, Z.~Zhu, N.~Lu, Y.~Cheng, and B.~Jiang, ``A data-driven degradation
  prognostic strategy for aero-engine under various operational conditions,''
  {\em Neurocomputing}, vol.~462, pp.~195--207, 2021.

\bibitem{t18}
A.~L. Ellefsen, E.~Bj{\o}rlykhaug, V.~{\AE}s{\o}y, S.~Ushakov, and H.~Zhang,
  ``Remaining useful life predictions for turbofan engine degradation using
  semi-supervised deep architecture,'' {\em Reliability Engineering \& System
  Safety}, vol.~183, pp.~240--251, 2019.

\bibitem{liu2019novel}
H.~Liu, Z.~Liu, W.~Jia, and X.~Lin, ``A novel deep learning-based
  encoder-decoder model for remaining useful life prediction,'' in {\em 2019
  International Joint Conference on Neural Networks (IJCNN)}, pp.~1--8, IEEE,
  2019.

\bibitem{t23}
Y.~Mo, Q.~Wu, X.~Li, and B.~Huang, ``Remaining useful life estimation via
  transformer encoder enhanced by a gated convolutional unit,'' {\em Journal of
  Intelligent Manufacturing}, vol.~32, pp.~1997--2006, 2021.

\bibitem{t5}
D.~Bahdanau, K.~Cho, and Y.~Bengio, ``Neural machine translation by jointly
  learning to align and translate,'' {\em arXiv preprint arXiv:1409.0473},
  2014.

\bibitem{t7}
D.~Xu, X.~Xiao, J.~Liu, and S.~Sui, ``Spatio-temporal degradation modeling and
  remaining useful life prediction under multiple operating conditions based on
  attention mechanism and deep learning,'' {\em Reliability Engineering \&
  System Safety}, vol.~229, p.~108886, 2023.

\bibitem{t8}
Y.-L. Liao, C.-C. Kuo, and Y.-F. Peng, ``Prediction and identification using
  recurrent wavelet-based cerebellar model articulation controller neural
  networks,'' in {\em The 2010 International Joint Conference on Neural
  Networks (IJCNN)}, pp.~1--6, IEEE, 2010.

\bibitem{t9}
Z.~Li, D.~Wu, C.~Hu, and J.~Terpenny, ``An ensemble learning-based prognostic
  approach with degradation-dependent weights for remaining useful life
  prediction,'' {\em Reliability Engineering \& System Safety}, vol.~184,
  pp.~110--122, 2019.

\bibitem{t12}
M.~Benker, L.~Furtner, T.~Semm, and M.~F. Zaeh, ``Utilizing uncertainty
  information in remaining useful life estimation via bayesian neural networks
  and hamiltonian monte carlo,'' {\em Journal of Manufacturing Systems},
  vol.~61, pp.~799--807, 2021.

\bibitem{t15}
C.~Zhang, P.~Lim, A.~K. Qin, and K.~C. Tan, ``Multiobjective deep belief
  networks ensemble for remaining useful life estimation in prognostics,'' {\em
  IEEE transactions on neural networks and learning systems}, vol.~28, no.~10,
  pp.~2306--2318, 2016.

\bibitem{t17}
H.~Dong, T.~Li, R.~Ding, and J.~Sun, ``A novel hybrid genetic algorithm with
  granular information for feature selection and optimization,'' {\em Applied
  Soft Computing}, vol.~65, pp.~33--46, 2018.

\bibitem{t20}
W.~Yu, I.~Y. Kim, and C.~Mechefske, ``An improved similarity-based prognostic
  algorithm for rul estimation using an rnn autoencoder scheme,'' {\em
  Reliability Engineering \& System Safety}, vol.~199, p.~106926, 2020.

\bibitem{t21}
H.~Cai, J.~Feng, W.~Li, Y.-M. Hsu, and J.~Lee, ``Similarity-based particle
  filter for remaining useful life prediction with enhanced performance,'' {\em
  Applied Soft Computing}, vol.~94, p.~106474, 2020.

\end{thebibliography}

\end{document}